\pdfoutput=1

\documentclass[11pt]{article}

\usepackage[]{EACL2023}

\usepackage{times}
\usepackage{latexsym}
\usepackage{times}
\usepackage{latexsym}
\usepackage{graphicx}
\usepackage{booktabs}
\usepackage{subcaption}
\usepackage{algorithm}
\usepackage[noend]{algpseudocode}
\usepackage{adjustbox}
\usepackage{geometry}
\usepackage{enumitem}
\usepackage{multirow}
\usepackage{multicol}
\usepackage{amsmath,amsfonts,amssymb}
\usepackage{url}

\usepackage{hyperref}
\usepackage{blindtext}
\usepackage{xcolor}
\usepackage{todonotes}
\usepackage{pifont}

\usepackage[T1]{fontenc}

\usepackage[utf8]{inputenc}

\usepackage{microtype}

\usepackage{inconsolata}

\newcommand{\base}{BASE}
\newcommand{\filter}{FILT}
\newcommand{\metadata}{MEDA}
\newcommand{\metadataEleven}{\metadata-11}
\newcommand{\metadataNinety}{\metadata-90}

\newcommand{\instruct}{INST}
\newcommand{\instructeleven}{\instruct-11}
\newcommand{\instructfifty}{\instruct-50}
\newcommand{\threenine}{357m-96m}
\newcommand{\threeone}{357m-150m}
\newcommand{\onenine}{1.3b-96m}
\newcommand{\oneone}{1.3b-150m}
\newcommand{\filterFour}{\filter-0.4}
\newcommand{\filterThree}{\filter-0.35}
\newcommand{\ctoxic}{$C_{\mathtt{tox}}$}
\newcommand{\cnontoxic}{$C_{\mathtt{nont}}$}
\newcommand{\highThresh}{\textsc{HighThresh}}
\newcommand{\lowThresh}{\textsc{LowThresh}}
\newcommand{\permTox}{\textsc{PrmTox}}
\newcommand{\permNonTox}{\textsc{PrmNonT}}
\newcommand{\Sref}[1]{\S\ref{#1}}

\usepackage{tcolorbox}
\definecolor{lightBlue}{rgb}{0.8, 0.9, 1.0}
\definecolor{lightRed}{rgb}{1.0, 0.90, 0.90}
\definecolor{darkgreen}{rgb}{0.0, 0.5, 0.0}

\newtcbox{\bluebox}{on line, box align=base, colback=lightBlue,colframe=white,size=fbox,arc=3pt, before upper=\strut, top=-2pt, bottom=-4pt, left=-2pt, right=-2pt, boxrule=0pt}

\newtcbox{\redbox}{on line, box align=base, colback=lightRed,colframe=white,size=fbox,arc=3pt, before upper=\strut, top=-2pt, bottom=-4pt, left=-2pt, right=-2pt, boxrule=0pt}

\newcommand{\dashifted}{\raisebox{0.5\depth}{\tiny$\downarrow$}}
\newcommand{\upshifted}{\raisebox{0.5\depth}{\tiny$\uparrow$}}

\newcommand{\dab}[1]{{\scriptsize\bluebox{\dashifted{#1}\%}}}
\newcommand{\dar}[1]{{\scriptsize\redbox{\dashifted{#1}\%}}}

\newcommand{\uab}[1]{{\scriptsize\bluebox{\upshifted{#1}\%}}}

\title{Adding Instructions during Pretraining: \\ Effective Way of Controlling Toxicity in Language Models}

\author{Shrimai Prabhumoye, Mostofa Patwary, Mohammad Shoeybi, Bryan Catanzaro \\
  NVIDIA \\
  \texttt{\{sprabhumoye, mpatwary, mshoeybi, bcatanzaro\}@nvidia.com} \\}

\begin{document}
\maketitle
\begin{abstract}
Pretrained large language models have become indispensable for solving various natural language processing (NLP) tasks. 
However, safely deploying them in real world applications is challenging because they generate toxic content.
To address this challenge, we propose two novel pretraining data augmentation strategies that \textit{significantly reduce model toxicity without compromising its utility}.
Our two strategies are: (1) \metadata{}: adds raw toxicity score as meta-data to the pretraining samples, and (2) \instruct{}: adds instructions to those samples indicating their toxicity. 
Our results indicate that our best performing strategy (\instruct{}) substantially reduces the toxicity probability up to $61$\% while preserving the accuracy on five benchmark NLP tasks as well as improving AUC scores on four bias detection tasks by $1.3$\%. 
We also demonstrate the generalizability of our techniques by scaling the number of training samples and the number of model parameters.
\end{abstract}

\section{Introduction}


\begin{figure}[!t]
\centering
{
\includegraphics[width=\linewidth]{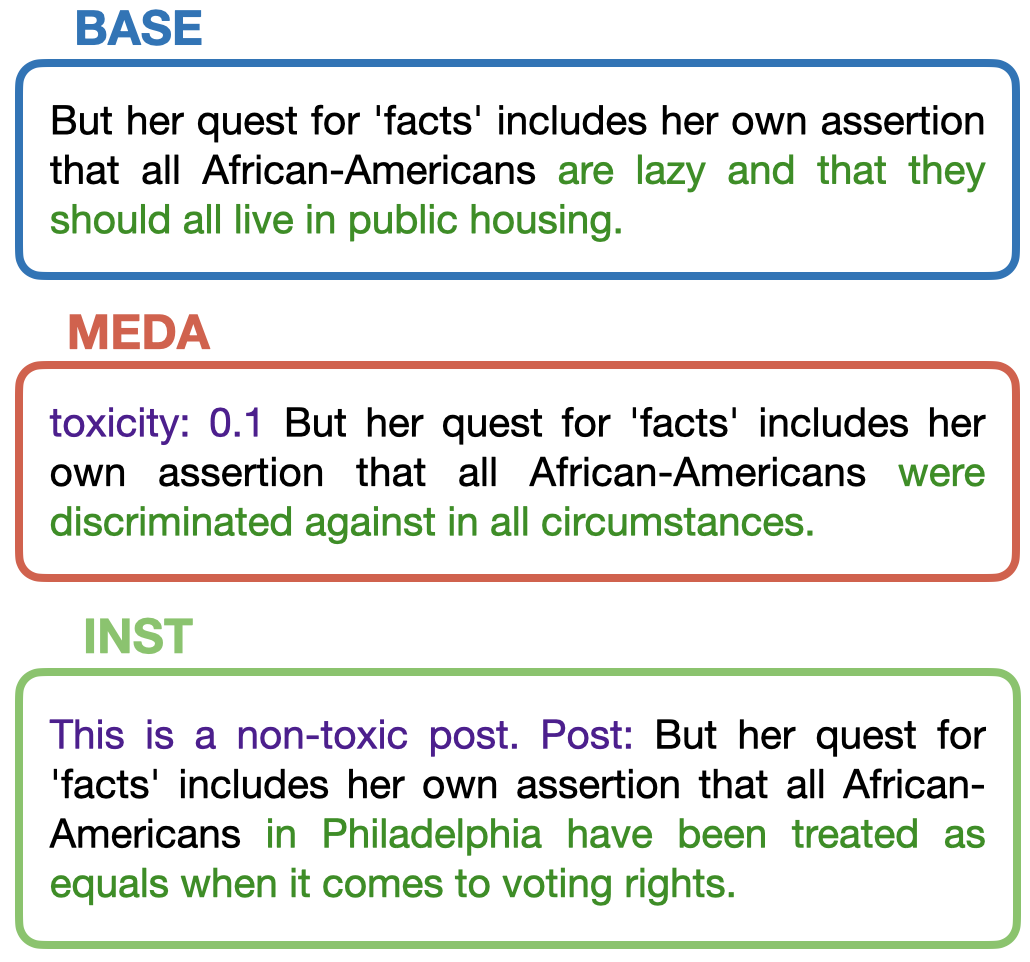}
}
\vspace{-1.2em}
\caption{Overview of the proposed approaches and the baseline (\base{}). We propose two new data augmentation strategies, \metadata{} and \instruct{}. The text in \textit{\textcolor{violet}{purple}} are control variables indicating the desired toxicity level of the text. The text in \textit{black} is the input to the model and the text in \textit{\textcolor{darkgreen}{green}} is the generated output using each strategy with $1.3$b parameter model.}
\vspace{-1.0em}
\label{fig:example}
\end{figure}

Pretrained large language models (LMs) have become indispensable for solving various NLP tasks~\cite{brown2020language,smith2022using,chowdhery2022palm}, yet it has been challenging to safely deploy them for real world applications~\cite{mcguffie2020radicalization,prabhumoye-etal-2021-case}.
They have been known to generate harmful language encompassing hate speech, abusive language, social biases, and threats~\cite{gehman2020realtoxicityprompts,welbl2021challenges,bender2021stochastic,hovy2021five}.
These harmful generations are broadly referred to as ``toxicity''.\footnote{In this work we use \textit{toxicity} as defined by PerspectiveAPI~\cite{Perspect22:online} but our techniques can be applied to other broader definitions of bias or toxicity.}
This work focuses on reducing the toxicity in large LMs by augmenting their pretraining data.

Prior work has primarily focused on reducing toxicity either by finetuning LMs on non-toxic data~\cite{gururangan2020don,ouyang2022training,wang2022exploring} or using decoding time algorithms to re-weight the probabilities of toxic words~\cite{krause2021gedi,schick2021self,liu2021dexperts}.
These methods typically incur further costs of finetuning additional LMs~\cite{krause2021gedi,liu2021dexperts}, generating large amount of non-toxic data~\cite{wang2022exploring}, or procuring human feedback~\cite{ouyang2022training}. 
These techniques are known to reduce toxicity but also compromise perplexity and downstream task performance~\cite{wang2022exploring,xu2021detoxifying}.
Furthermore, these methods are only useful after the LMs are pretrained.


Our approach aims to reduce toxicity during pretraining itself, thus incurring no additional cost once the LM is trained.
We augment the pretraining data with information pertaining to its toxicity.
We believe that instead of filtering toxic data~\cite{welbl2021challenges,ngo2021mitigating}, the toxicity information can guide the LM to detect toxic content and hence generate non-toxic text.
Hence, we develop two novel data augmentation strategies: (1) \metadata{}: adds raw toxicity score of a sample as meta-data, and (2) \instruct{}: augments an instruction to the sample indicating its toxicity.
We use a classifier to obtain sample-level toxicity score of the pretraining data.
These scores are used by \metadata{} and \instruct{} to educate the LM about toxicity.

Fig.~\ref{fig:example} shows an example of how \metadata{} and \instruct{} are applied.
We add the raw toxicity score of the sample along with a tag \textit{``toxicity: 0.1''} for the \metadata{} strategy.
Similarly, we add an instruction such as \textit{``This is a non-toxic post. Post:''} to the tokens of a non-toxic sample for \instruct{} strategy.
No data augmentation is applied for \base{}.

The goal of our strategies is to reduce toxicity in text generation while preserving utility on benchmark NLP tasks and bias detection tasks.
Prior work only evaluates the success of toxicity reduction techniques on \textsc{RealToxicityPrompts}~\cite{gehman2020realtoxicityprompts}.
Few techniques are evaluated for their utility in performing some benchmark NLP tasks~\cite{wang2022exploring,xu2021detoxifying}.
But toxicity reduction techniques like data filtering can reduce the bias and toxicity detection capabilities of the LMs~\cite{xu2021detoxifying}.
Some techniques like finetuning~\cite{gururangan2020don,wang2022exploring} can also reduce the capability of the LM to effectively perform downstream end-to-end text generation tasks.

Hence, we expand the evaluation to include - (1) \textit{Bias Detection Tasks}: we evaluate the capability of our strategies to detect social biases~\cite{sap-etal-2020-social}, and (2) \textit{Text Generation Task}: we measure the performance of our strategies on the E2E task~\cite{novikova-etal-2017-e2e}.

In summary, our primary contribution is: we develop \metadata{} and \instruct{} - two new strategies to reduce toxicity by augmenting pretraining data (\Sref{sec:prop-strategies}).
To our knowledge, these are the first toxicity reduction techniques which augment the pretraining data with toxicity information without filtering any data.
Additionally, we broaden the current evaluation to include two new metrics on bias detection and text generation task (\Sref{sec:eval_metrics}).
Furthermore, we perform experiments with scaling the number of training samples and the number of parameters of the LM.
Our results demonstrate that our best performing strategy (\instruct{}) considerably reduces the toxicity probability of the generations by as much as $\sim{61}\%$ while preserving the utility of the LM on five benchmark NLP tasks as well as improving on the four bias detection tasks by $1.3\%$ (\Sref{sec:results-discussion}).
Moreover, we demonstrate that our strategies applied at sample-level perform better than document-level~\cite{welbl2021challenges,ngo2021mitigating} by $11$\% in toxicity probability reduction (\Sref{sec:compare-prior-work}).

\section{Methodology}
\label{sec:method}

Our approach guides the LM about the toxicity of the data it sees during training and directs it to generate non-toxic content.
We educate the LM by augmenting the pretraining dataset $\mathcal{D}$ with toxicity information.
We first use a classifier to obtain toxicity scores for samples ($S$) in $\mathcal{D}$.
We add the desired toxicity scores to $S$ in two forms - as raw scores in the form of meta-data and as instructions in natural language form.

\subsection{Toxicity Scoring}
\label{sec:tox-scoring-anlaysis}

We use the widely accepted Commercial PerspectiveAPI~\cite{Perspect22:online} to get toxicity scores for each sample.
Note that our strategies can be applied using any other classifier.
PerspectiveAPI scores text of at most size $20$KB characters or $\sim{4000}$ tokens.
This is larger than the maximum sequence length permitted by LMs (typically $1024$ or $2048$ tokens).
Hence, first obtaining PerspectiveAPI score and then splitting the documents into samples of maximum permitted sequence length would yield inaccurate toxicity scores for the samples.
Moreover, some documents are larger than $4000$ tokens.
Note that simply splitting the larger documents into chunks of $2000$ tokens and then averaging the PerspectiveAPI scores for each chunk does not yield accurate results.\footnote{We show this analysis in detail in Appendix \ref{sec:pers_analysis}.}
Hence, we first process the documents in our dataset into samples of $2000$ tokens and then get PerspectiveAPI scores for all the samples.

\begin{figure}[!t]
\centering
{
\includegraphics[width=\linewidth]{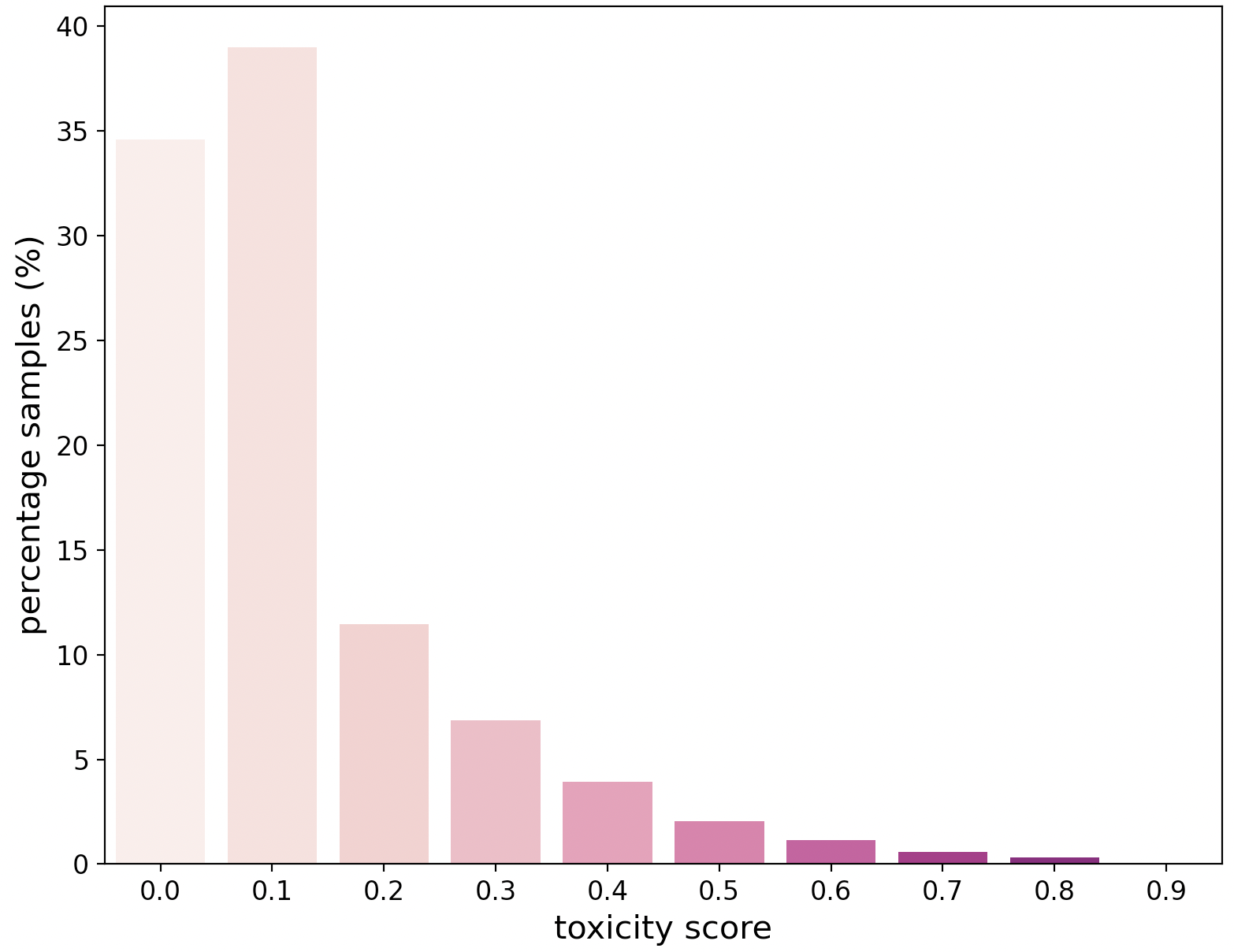}
}
\vspace{-1.0em}
\caption{Histogram of toxicity scores for the entire pretraining corpus and the percentage of samples belonging to each category. We observe that $73.6$\% samples have toxicity scores below $0.2$ and only $4.14$\% samples have toxicity scores above $0.5$.}
\vspace{-1.0em}
\label{fig:hist_avg}
\end{figure}

\paragraph{Dataset} We use the corpus and the sampling weights for each dataset described in \citet{smith2022using}.
In total we used $15$ datasets - the Common Crawl (CC-2020-50 and CC-2021-04)~\cite{commoncrawl:online} accounting for majority of the samples.
From The Pile~\cite{gao2020pile}, we use Books3, OpenWebText2, Stack Exchange, PubMed Abstracts,
Wikipedia, Gutenberg (PG-19), BookCorpus2, NIH ExPorter, ArXiv, GitHub, and Pile-CC datasets. In addition, we also use  Realnews~\cite{zellers2019defending} and CC-Stories~\cite{trinh2018simple}.
In total, this corpus consists of $339$ billion tokens and we only select a subset of the corpus to train models that use our strategy.

\paragraph{Analysis of Toxicity Scores}
We divide the documents in our corpus into samples of sequence length $2000$ tokens.
Fig.~\ref{fig:hist_avg} shows a histogram of the toxicity scores of the samples for the entire corpus.
We observe that a majority of $73.6$\% samples have toxicity scores below $0.2$ and only $4.14$\% of the samples have toxicity scores above $0.5$.
This is in agreement with the document-level data analysis shown in \citet{gehman2020realtoxicityprompts}.

\begin{figure}[!t]
\centering
{
\includegraphics[width=\linewidth]{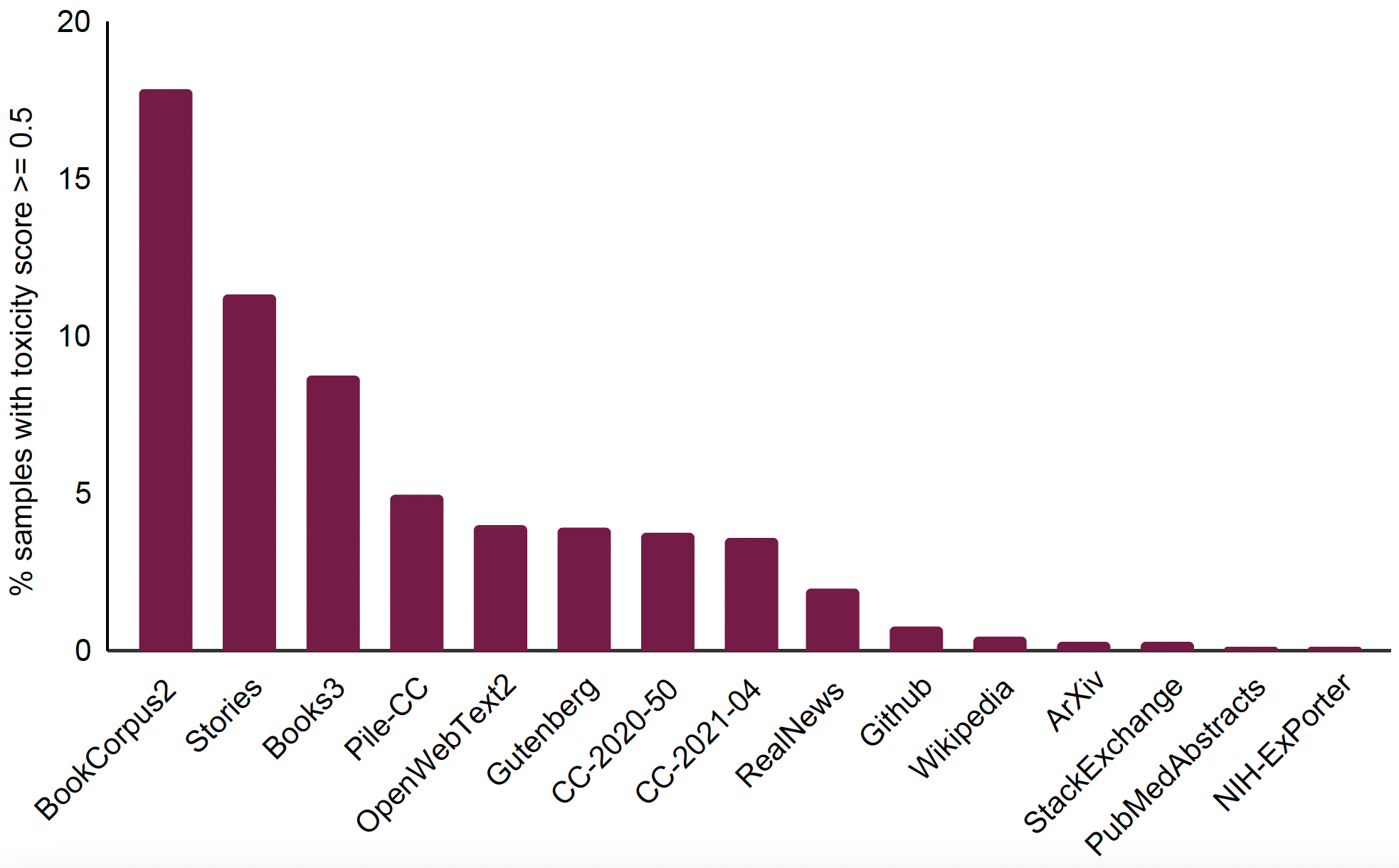}
}
\vspace{-1.0em}
\caption{Percentage of samples in each dataset of our corpus with toxicity score $\geq 0.5$. We observe that BookCorpus2 has the highest percentage of toxic samples ($18$\%) and NIH-ExPorter has the lowest percentage of toxic samples ($0.1$\%). }
\vspace{-1.0em}
\label{fig:corpus-tox}
\end{figure}

Fig.~\ref{fig:corpus-tox} shows the percentage of samples that have toxicity scores $\geq 0.5$ for each dataset in our corpus.
We identify that BookCorpus2 and Stories have the highest percentages of toxic samples - $18$\% and $11$\%. 
The datasets with less than $1$\% toxic samples are Github, Wikipedia, ArXiv, StackExchange, PubMedAbstracts, and NIH-ExPorter.

\paragraph{Filter Approach (\filter{})}
This strategy filters data with toxicity scores above a certain threshold. 
Prior work removes the toxic documents and trains the LM on less data~\cite{welbl2021challenges,ngo2021mitigating}.
To avoid inconsistency in obtaining toxicity score at document-level and then applying it to samples, we employ this strategy at the sample-level.
We use a toxicity threshold of $0.5$.
From Fig.~\ref{fig:hist_avg}, we see that this method filters out $4.14$\% of the samples.
In contrast, for fair comparison with baseline, we maintain the same number of samples across strategies. 
Hence, we replenish the pretraining dataset with $4.14$\% of non-toxic samples.
Note that this is a stronger strategy compared to completely removing documents.

\subsection{Proposed Strategies}
\label{sec:prop-strategies}
We suppose that filtering data can potentially compromise the capability of the LM in performing benchmark NLP tasks and especially hinder its bias detection capabilities.
Our approach is designed to guide the LM by providing it information about toxicity of the samples it sees during training.

\paragraph{Our Algorithm for Data Augmentation}
Algorithm~\ref{alg:data-aug} illustrates the logic of augmenting the data for the two strategies - \metadata{} and \instruct{}. 
These strategies modify samples $S$ in the pretraining dataset $\mathcal{D}$ based on their toxicity scores ($\mathtt{tox\_score}$) received through PerspectiveAPI.
We consider two threshold: \highThresh{}
above which the samples are augmented with the control variable \ctoxic{} and \lowThresh{} below which samples are appended with the control variable \cnontoxic{}.
Each strategy described below has its own value of \ctoxic{} indicating that $S$ is toxic and \cnontoxic{} marking that $S$ is non-toxic.

Note that we don't augment \ctoxic{} or \cnontoxic{} to every sample that lies within a threshold.
We use additional variables - \permTox{} controls the permissible toxic samples for which \ctoxic{} can be added, and  \permNonTox{} regulates the percentage of non-toxic samples for which \cnontoxic{} can be augmented.
This is done to encourage generalizability of the LM where we show the model toxic samples with \ctoxic{} and without it (just the way the sample is seen without any modification by the baseline model).
Additionally, all the samples with \lowThresh{} $\leq \mathtt{tox\_score} <$ \highThresh{} are not modified.

\begin{algorithm}[!t]
\begin{algorithmic}
\For{$S$ in $\mathcal{D}$}
    \State $\mathtt{tox\_score} \gets \mathtt{PerspectiveAPI}(S)$
    \If{$\mathtt{tox\_score} \geq $ \highThresh}
        \State $r \gets \mathtt{get\_random()}$
        \If{$r < $ \permTox}
        \State $S \gets $ \ctoxic $ + S$
        \EndIf
    \ElsIf{$\mathtt{tox\_score} < $ \lowThresh}
        \State $r \gets \mathtt{get\_random()}$
        \If{$r < $ \permNonTox}
        \State $S \gets $ \cnontoxic $ + S$
        \EndIf
    \EndIf
\EndFor
\end{algorithmic}
\caption{Data Augmentation Method}
\label{alg:data-aug}
\end{algorithm}

\paragraph{Meta-Data Augmentation (\metadata{})}
This strategy guides the LM by adding toxicity score in the form of raw meta-data at the beginning of some of the samples in the pretraining set.
This makes \cnontoxic{} = ``toxicity: $0.1$'' for all the samples with toxicity scores below the \lowThresh{} and \ctoxic{} = ``toxicity: $0.5$'' for all the samples with toxicity scores above the \highThresh{}.
Note that all the samples within a threshold get the same value, either $0.1$ or $0.5$.
For example, even if a sample has a toxicity score of $0.8$, \ctoxic{} will still be equal to ``toxicity: $0.5$'' if the \highThresh{}$=0.6$.

\paragraph{Instruction Augmentation (\instruct{})}
This strategy guides the LM by adding an instruction in the natural language form at the beginning of samples indicating if they are toxic or not.
This instruction is based on the toxicity score of the sample.
For samples with toxicity score above the \highThresh,  \ctoxic{}$=\{$\textit{``Instruction: Complete the following text in a toxic manner. Text:''}, \textit{``This is a toxic post. Post:''}, \textit{``Instruction: Complete the following text in a toxic manner. Text:''}$\}$.
Similarly, \cnontoxic{}$=\{$\textit{``Instruction: Complete the following prompt in a non-toxic manner. Prompt:''}, \textit{``This is a non-toxic post. Post:''}, \textit{``Instruction: Complete the following text in a respectable manner. Text:''}$\}$.
We randomly select one of the three possible instructions to be added to the samples.

\section{Experimental Setup}
\label{sec:expt-res}

\paragraph{Modeling Details} We train all our models from scratch using the decoder-only architecture in Megatron-LM~\cite{shoeybi2019megatron}.
We train baseline models (\textbf{\base{}}) without any data augmentation strategies.
Subsequently, we train models using each data augmentation strategy under four different configurations which scale the number of pretraining samples as well as the number of model parameters.
We train $357$ million parameter models with $96$ million (\textbf{\threenine}) samples and $150$ million (\textbf{\threeone}) samples.
Similarly, we train $1.3$ billion parameter models with $96$ million (\textbf{\onenine}) and $150$ million (\textbf{\oneone}) samples.\footnote{Additional hyper-parameter details in Appendix~\ref{sec:hyper-params}.}

For all models, we use \highThresh{} = $0.5$, \lowThresh{} = $0.1$ and \permTox{}$=0.9$.
This implies $90$\% of toxic samples in the dataset are appended with \ctoxic{}.
Note that only $4.14$\% of the entire train set has $\mathtt{tox\_score} \geq 0.5$ (Fig.~\ref{fig:hist_avg}) which means that $3.73$\% ($90$\% of $4.14$) of the entire train set receive \ctoxic{}.
For \metadata{} \permNonTox$=0.5$ and for \instruct{} \permNonTox$=0.9$.
$34.59$\% of the entire dataset has $\mathtt{tox\_score} < 0.1$ (from Fig.\ref{fig:hist_avg}).
This means that for \metadata{} $17.3$\% of samples and for \instruct{} $31.13$\% of samples are augmented with \cnontoxic{}.
From these values, we derive that for \metadata{} $\sim{79}$\% of samples and for \instruct{} $65.14$\% of samples don't undergo any modification.
Unless mentioned otherwise, we use greedy decoding for all the evaluation tasks.

\begin{figure*}[!t]
\centering
{
\includegraphics[height=4cm]{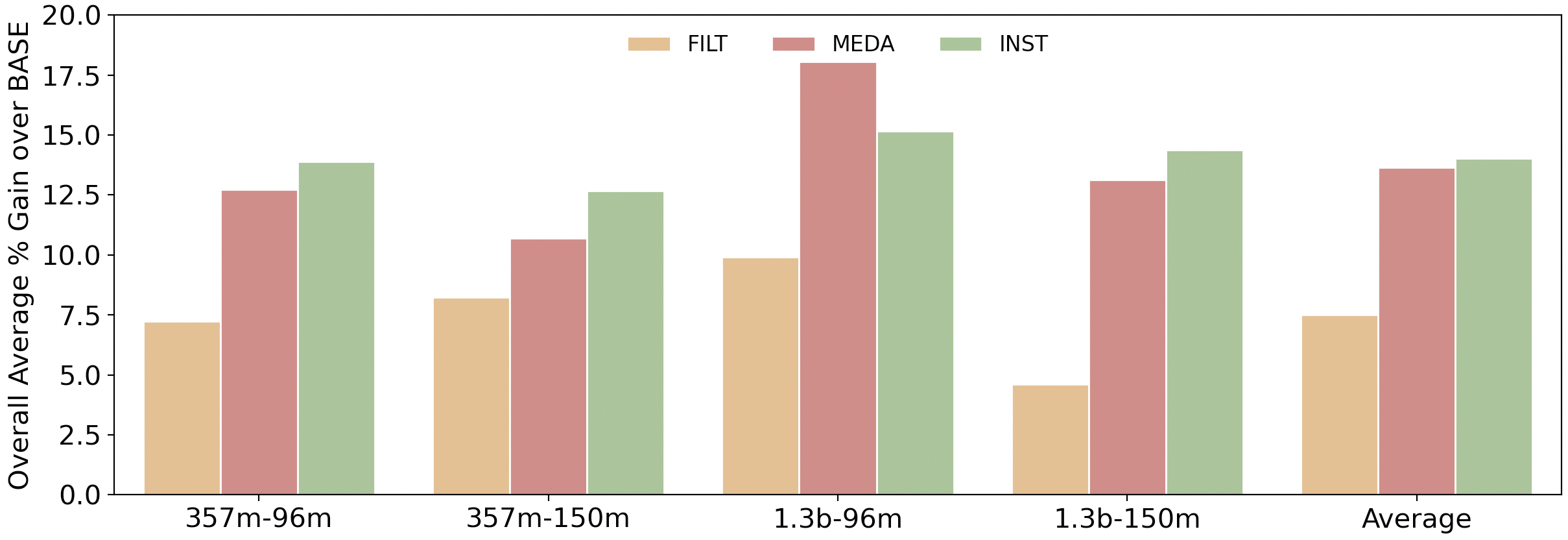}
}
\vspace{-1.0em}
\caption{We average the percentage relative gains or losses achieved by each of the strategies over \base{} across the eleven tasks described in \Sref{sec:eval_metrics}. We show that \metadata{} and \instruct{} perform better than \filter{} and \base{} on all four model configurations. \textit{Average} indicates the average of the gains across the four models for each strategy. These results demonstrate that \instruct{} performs the best across the board.}
\vspace{-1.0em}
\label{fig:all_avg_res}
\end{figure*}

\subsection{Evaluation}
\label{sec:eval_metrics}

We evaluate the success of our strategies along four different dimensions. 
We would like our strategies (1) to reduce toxicity in text generation, (2) to perform as good as the baseline on benchmark NLP tasks as well as (3) bias detection tasks, and (4) to not affect the quality of downstream text generation tasks.
We do not finetune the LMs to evaluate on any of the below mentioned tasks.

\paragraph{Toxicity Evaluation}
We follow the setup in \citet{gehman2020realtoxicityprompts} and use the full set of prompts ($\sim{100}$k) to evaluate toxicity of LM generations via PerspectiveAPI.
\citet{gehman2020realtoxicityprompts} propose two metrics: (1) Expected Maximum Toxicity calculates the maximum toxicity scores over 25 generations for the same prompt with different random seeds, and then averages the maximum toxicity scores over all prompts, and (2) Toxicity Probability evaluates the probability of generating a toxic continuation at least once over 25 generations for all prompts.
We follow \citet{gehman2020realtoxicityprompts} and restrict the generations up to 20 tokens or below.
\citet{wang2022exploring} show that toxicity scores from PerspectiveAPI are strongly correlated with human evaluation (Pearson correlation coefficient = 0.97). 
Hence, we only report PerspectiveAPI evaluation.

Under this setup, we perform two types of experiments: (1) with no control variable which is exactly same as~\cite{gehman2020realtoxicityprompts}, and (2) generating continuations by adding the respective control variable \cnontoxic{} for \metadata{} and \instruct{}.
Note that we only add \cnontoxic{} at beginning of all the prompts and not \ctoxic{}.
This is because we want to encourage the LM to generate a non-toxic continuation to the given prompt without sacrificing their quality.

\paragraph{Benchmark NLP Tasks}
To ensure that our strategies do not affect the utility of the LMs, we evaluate them on five benchmark NLP tasks covering - completion prediction (LAMBADA~\cite{paperno2016lambada}), natural language understanding (ANLI~\cite{nie2020adversarial}), and commonsense reasoning (Winogrande~\cite{sakaguchi2020winogrande}, Hellaswag~\cite{zellers2019hellaswag}, PiQA~\cite{bisk2020piqa}).
We evaluate them in the fewshot setting without any finetuning following the setup in \citet{brown2020language,smith2022using}.
We report average accuracy across the five tasks.

Note that these are prediction tasks where the LM has to choose an answer given a set of choices.
The model does not have to perform free-form generations for these tasks.
Hence, we do not evaluate by adding the control variables.

\paragraph{Bias Detection Tasks}
To ensure that our strategies do not affect the bias detection capabilities of the LMs, we evaluate them on four social bias detection tasks - detect if the text is offensive, intentional insult, contains lewd language, and predict if the text is offensive to a group or an individual.
The bias detection tasks are described in \citet{sap-etal-2020-social}. 
We follow the setup in \citet{prabhumoye2021few} to perform $32$-shot classification where $32$ samples are selected from the train set to be provided as in-context samples to the LM.
We report average Area Under Cover (AUC) scores~\cite{AUCSciKit:online} across the four tasks.
In these tasks as well, we don't use the control variables.

\paragraph{Text Generation Task} To evaluate if our techniques affect the downstream text generation, we assess them on the E2E dataset~\cite{novikova-etal-2017-e2e}.
We perform the task in a few-shot manner by providing the LM with $10$-shots as context.
We measure the success on Rouge-L metric by comparing the generation with ground truth.
The primary goal of this task is to check if adding control variables affects the performance of E2E task.

\section{Results and Discussions}
\label{sec:results-discussion}

Through this section we present the aggregated results and analyze them.
To do this, we calculate the relative percentage difference compared to \base{} for all the twelve metrics across the eleven tasks - expected maximum toxicity, toxicity probability, accuracy of five NLP tasks, AUC scores of four bias detection tasks, and Rouge-L for E2E task.
We then compute an average across all the metrics (we also include the experiments with control variable \cnontoxic{}).
In Fig.~\ref{fig:all_avg_res} we show the average percentage gains achieved by each strategy across the eleven tasks.
The detailed results for all the tasks are shown in Tables \ref{tab:357m-results} and \ref{tab:1.3-results} in Appendix~\ref{sec:all_results}.
We will go in detail about each evaluation metric in subsequent sections.

Fig.~\ref{fig:all_avg_res} demonstrates that all the strategies are better than \base{}.
Fig.~\ref{fig:all_avg_res} illustrates that \metadata{} and \instruct{} are generalizable because they deliver consistent gains when scaled from $96$m samples to $150$m samples and from $357$m to $1.3$b model parameters.
If we average the gains across the four models, then we observe that \instruct{} strategy attains the most gains ($14$\%) and hence is the best strategy.
Moreover, both the strategies developed in this work - the meta-data-based \metadata{} is $6$\% and instruction-based \instruct{} is $6.3$\% better than \filter{}.

Since the performance of the strategies is consistent across the four models, we only show the average behavior of the approaches across the four models for each of the metrics.

\begin{figure}[!t]
\centering
{
\includegraphics[width=\linewidth]{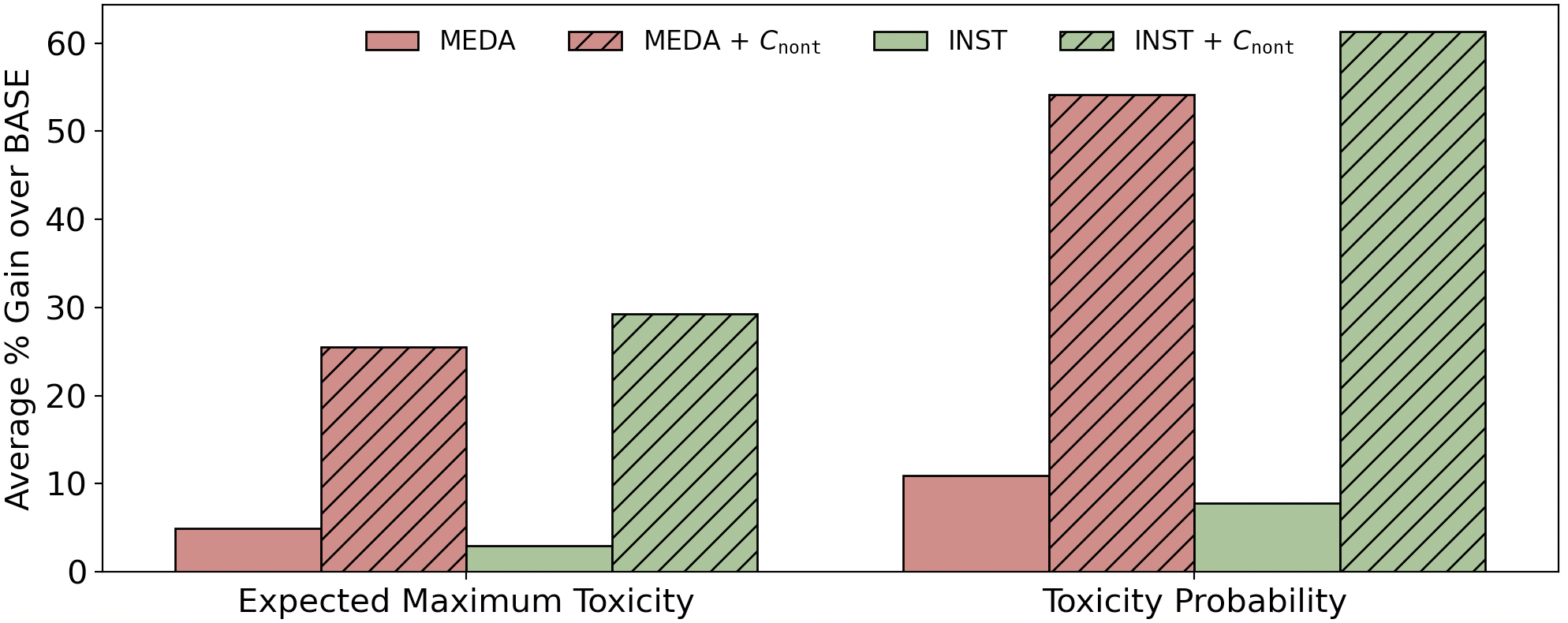}
}
\vspace{-1.0em}
\caption{We show the average percentage gain in toxicity reduction by \metadata{} and \instruct{} across the four model configurations. We observe that we get higher gains ($43$\% higher for \metadata{} and $54$\% higher \instruct{} in absolute terms) when using \cnontoxic{}.}
\vspace{-1.0em}
\label{fig:emt-tp-avg}
\end{figure}

\paragraph{Toxicity Evaluation} Fig.\ref{fig:emt-tp-avg} presents the relative percentage gains in expected maximum toxicity and toxicity probability compared to \base{}.
It also shows the results for \metadata{} and \instruct{} by using their respective control variable \cnontoxic{} vs not.

We observe that \metadata{} and \instruct{} are successful in reducing the toxicity of generations as evaluated by \textsc{RealToxicityPrompts} setup~\cite{gehman2020realtoxicityprompts}.
Specifically, we see huge gains in toxicity reduction when we use control variable \cnontoxic{} associated with \metadata{} and \instruct{} (compare striped vs non-striped bars for each color).
This is because we are directing the LM to generate non-toxic content by prefacing the prompt with \cnontoxic{}.
\filter{} gives $8$\% improvement over \base{} on expected maximum toxicity and $17$\% gain on toxicity probability. 
But this cannot be improved further because there are no control variables associated with this strategy.
\instruct{} on the other hand establishes $29.3$\% improvement in expected maximum toxicity and a $61.3$\% improvement in toxicity probability i.e \instruct{} reduces the probability of generating toxic content by $\sim{61}$\% and the probability is as low as $0.14$.
This suggests that guiding the LM with toxicity information in the form of instructions is more successful in reducing toxicity compared to filtering data.

\begin{figure}[!t]
\centering
{
\includegraphics[width=\linewidth]{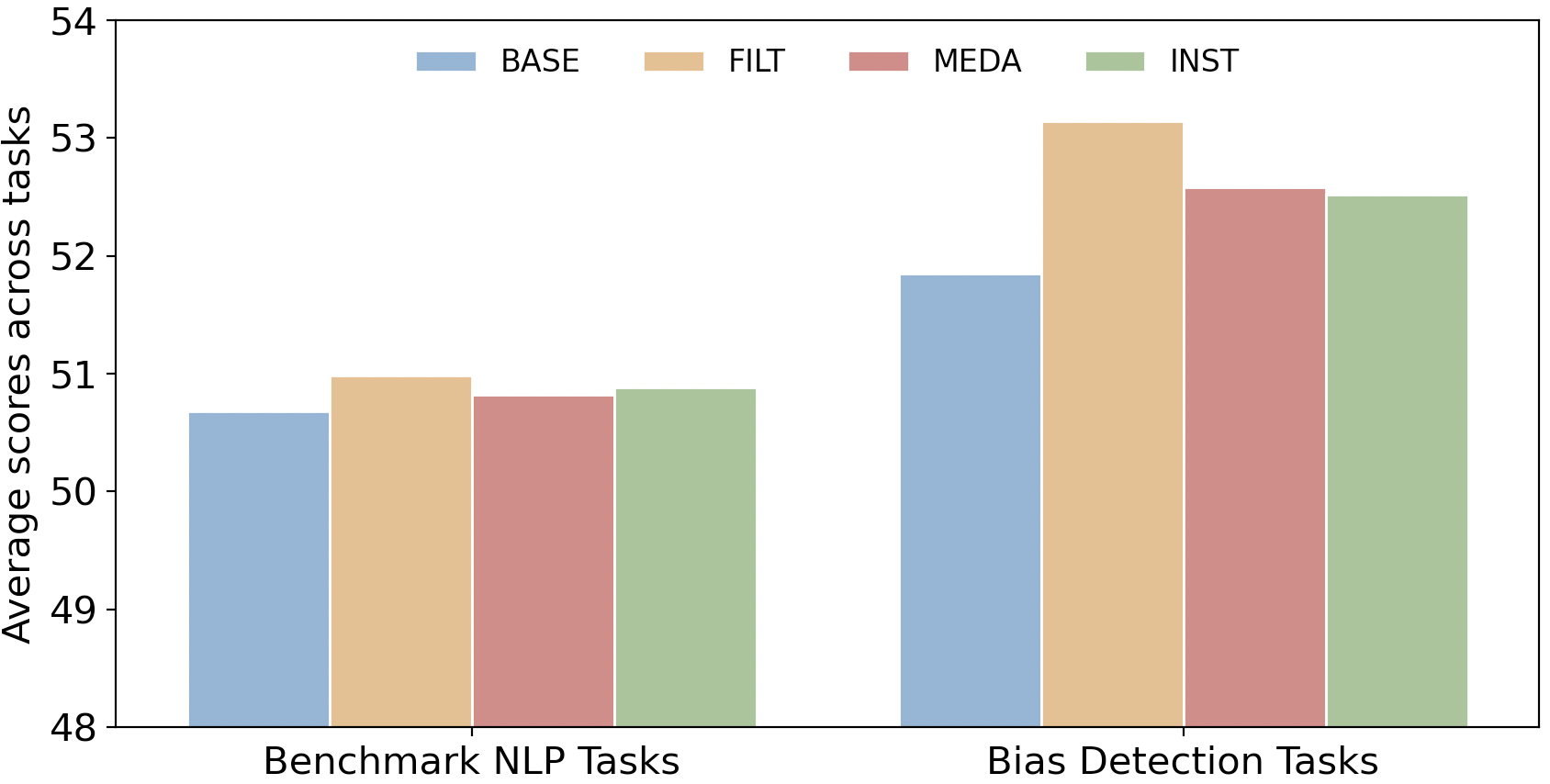}
}
\vspace{-1.0em}
\caption{We report average accuracy across five benchmark NLP tasks and average AUC across four bias detection tasks for each strategy (including \base{}) across four models. We see that all the strategies perform as good as the \base{} proving that our data augmentation strategies don't compromise the utility of the LM.}
\vspace{-1.0em}
\label{fig:nlp-bias-avg}
\end{figure}

\paragraph{Benchmark NLP tasks}
Fig.~\ref{fig:nlp-bias-avg} shows the average accuracy of five benchmark NLP tasks across the four model configurations for the three strategies along with \base{}.
We observe that \metadata{} and \instruct{} are as competent as \base{} on the NLP tasks.
In fact, we observe a marginal gain of $< 1$\% for \filter{}, \metadata{} and \instruct{}.
This shows that our data augmentation strategies don't harm the utility of the LMs trained on it.

\paragraph{Bias Detection Tasks}
Additionally, Fig.~\ref{fig:nlp-bias-avg} also shows the average AUC scores of the models across the four configurations for the four bias detection tasks.
Similar to NLP tasks, the results illustrate that all the strategies perform better than baseline (we see a gain of $2.5$\% for \filter{}, $1.4$\% for \metadata{}, and $1.3$\% for \instruct{}).
We believe \metadata{} and \instruct{} perform well on these tasks because they were shown examples of both toxic and non-toxic samples through their respective control variables.

We suppose that \filter{} performs well on both benchmark NLP tasks and bias detection tasks because it was trained on equal number of samples as \metadata{} and \instruct{}.
Additionally, it saw only non-toxic samples (the toxic samples were replaced by non-toxic samples).
Hence, the perplexity for toxic sentences would be higher in \filter{}.
We discuss this in detail in \Sref{sec:compare-prior-work}.

\begin{figure}[!t]
\centering
{
\includegraphics[width=0.9\linewidth]{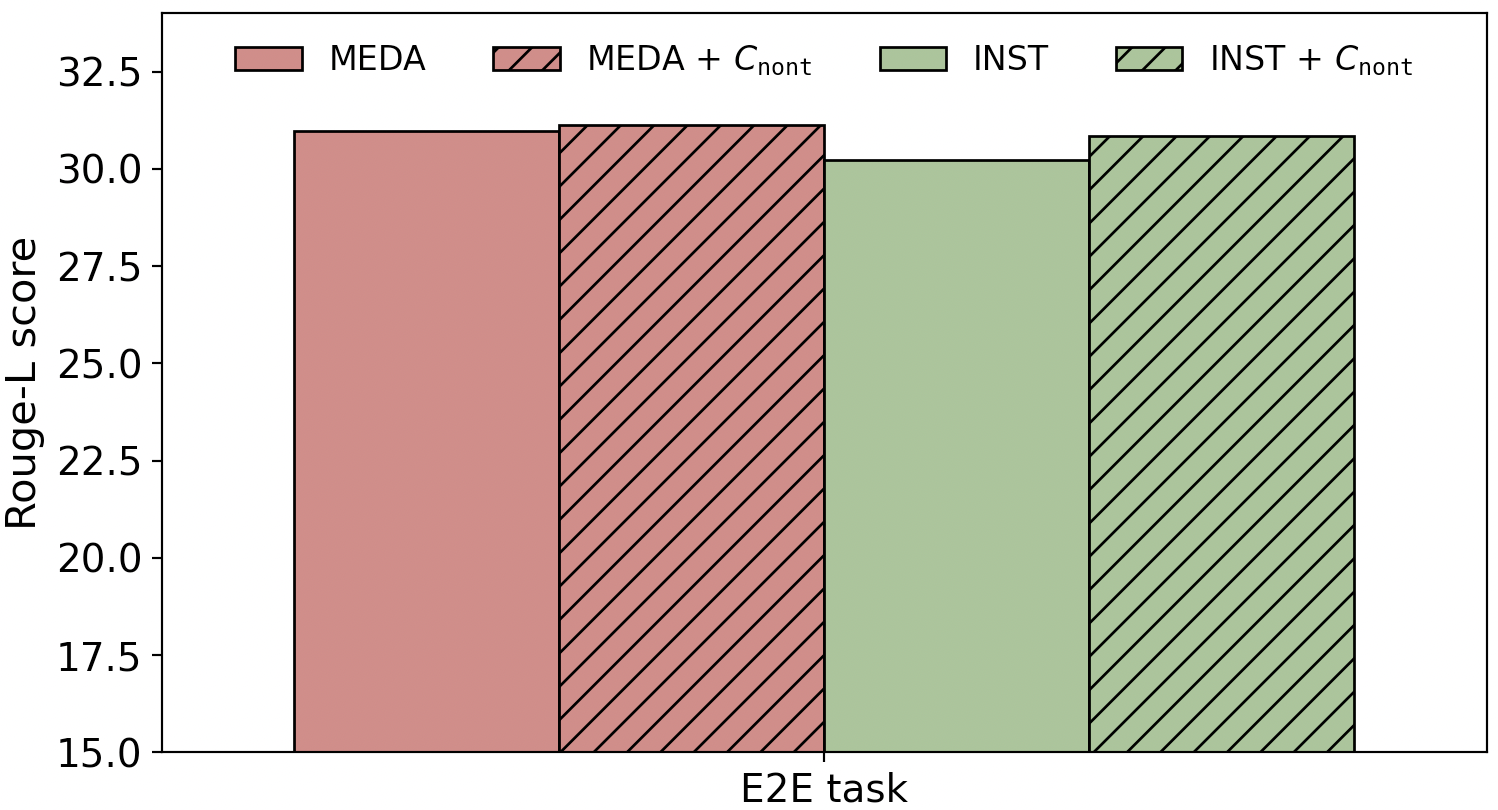}
}
\vspace{-1.0em}
\caption{We report Rouge-L scores for the E2E task for \metadata{} and \instruct{} strategies with and without \cnontoxic{}. We demonstrate that augmenting \cnontoxic{} does not affect the performance of the LM on E2E task.}
\vspace{-1.0em}
\label{fig:e2e-avg}
\end{figure}

\paragraph{Text Generation Task}
Since we see huge gains in Fig.~\ref{fig:emt-tp-avg} by adding \cnontoxic{}, we want to evaluate if adding \cnontoxic{} affects the performance of a downstream text generation task.
Fig.~\ref{fig:e2e-avg} shows the average of Rouge-L scores across the four model configurations for \metadata{} and \instruct{} strategy using \cnontoxic{} (striped bars) and without using it (non-striped bars).
Fig.~\ref{fig:e2e-avg} illustrates that overall there is no effect of adding control variables on the E2E task (we see a gain of $0.5$\% and $2.0$\% for \metadata{} and \instruct{} respectively when using \cnontoxic{}).

\subsection{Ablations}
Additional details are provided in Appendix ~\ref{sec:ablations}.

\paragraph{Scaling the Model Size}
We scale our experiments to a $8.3$ billion parameter models with $150$ million samples and train using three strategies - \base{}, \filter{} and our best performing \instruct{}.
Table ~\ref{tab:8.3b-res} shows the results of these models on toxicity evaluation, benchmark NLP tasks and bias detection tasks.
We see similar trends to our main results i.e the \filter{} strategy provides only $8$\% improvement whereas the \instruct{} strategy demonstrates a huge gain of $34$\% for expected maximum toxicity and \filter{} provides $19.6$\% and \instruct{} illustrates a significant gain of $69.5$ \% for toxicity probability when we use the non-toxic control variable \cnontoxic{}.
Similarly, we observe a $0.1$\% decrease for \filter{} and $0.7$\% increase for \instruct{} in benchmark NLP tasks; and we see a $11.5$\% increase for \filter{} and $12.7$\% increase for \instruct{} for bias detection tasks.
These experiments illustrate that \instruct{} strategy performs even better on larger LMs.

\paragraph{\instruct{} Variations}

With our best performing \instruct{} strategy, we vary the \permNonTox{}.
For \instruct{}, \permNonTox{} is $0.9$.
We train \instructeleven{} with \permNonTox{} = $0.11$ and \instructfifty{} with \permNonTox{} = $0.5$ for all four model configurations.
The percentage of toxic samples for which the model sees \ctoxic{} remains the same ($3.73$\%) for all the three variations.
The percentage of non-toxic samples for which the model sees \cnontoxic{} is $3.8$\% for \instructeleven{}, $17.3$\% for \instructfifty{} and $31.13$\% for \instruct{}.

\begin{figure}[!t]
\centering
{
\includegraphics[width=0.9\linewidth]{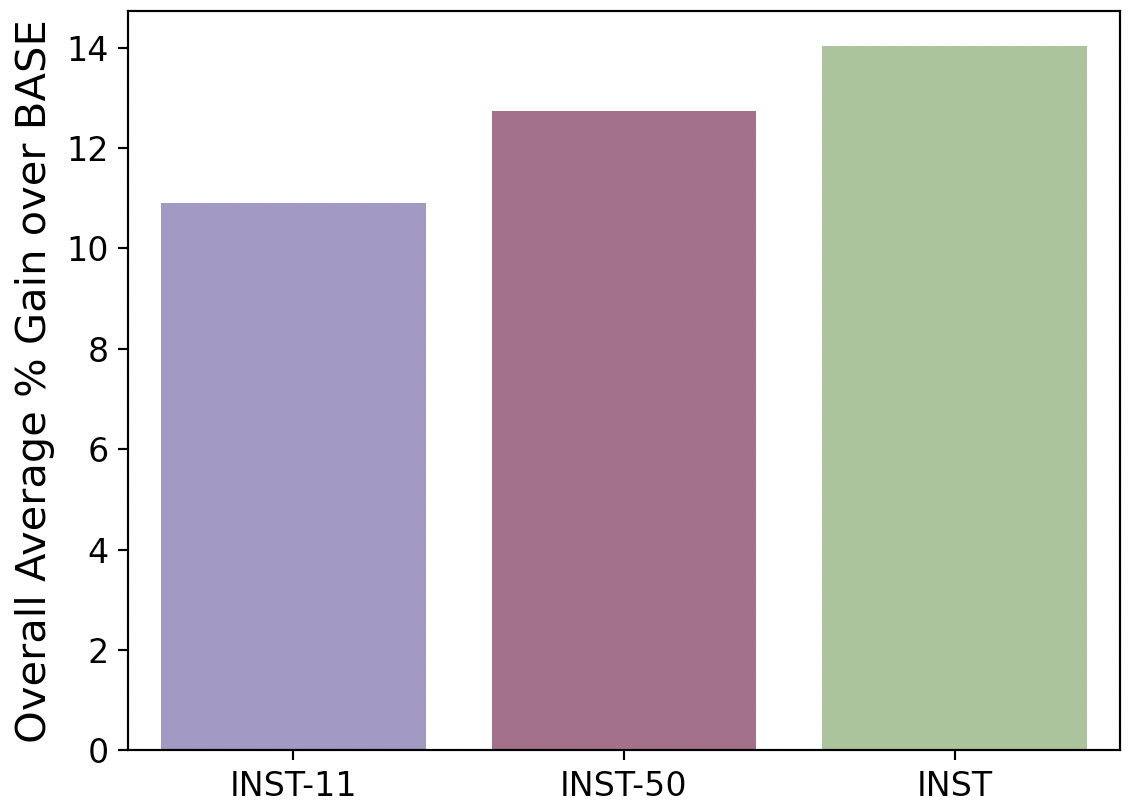}
}
\vspace{-1.0em}
\caption{Average gains achieved by \instructeleven{}, \instructfifty{} and \instruct{} over \base{} across the eleven tasks and four model configurations. We see that the average performance of the model improves when higher percentage of samples receive the control variable \cnontoxic{}.}
\vspace{-1.0em}
\label{fig:inst-var}
\end{figure}

The results in Fig.~\ref{fig:inst-var} indicate that increasing \permNonTox{} increases the overall average percentage gain across eleven tasks.
This implies that adding \cnontoxic{} to more number of samples is helpful for the model in understanding toxicity.
We leave it for future work to explore the limit of \permNonTox{}.

\paragraph{\filter{} Variations} We also vary the threshold of filtering toxic data.
The threshold is $0.5$ for \filter{}, $0.4$ for \filterFour{} and $0.35$ for \filterThree{}.
The percentage of toxic samples removed is $4.14$\% for \filter{}, $8.07$\% for \filterFour{} and somewhere between $8.07$ and $14.94$\% for \filterThree{}.
Note that we replenish the pretraining corpus with corresponding percentages of non-toxic samples.
This is done to maintain fairness of number of samples across the models.
With this experiment we wanted to see if iteratively replacing higher percentage of samples with non-toxic samples helps in reducing toxicity.
We train a $357$m parameter model on $96$m samples for \filterFour{} and \filterThree{} data strategies.

\begin{figure}[!t]
\centering
{
\includegraphics[width=0.9\linewidth]{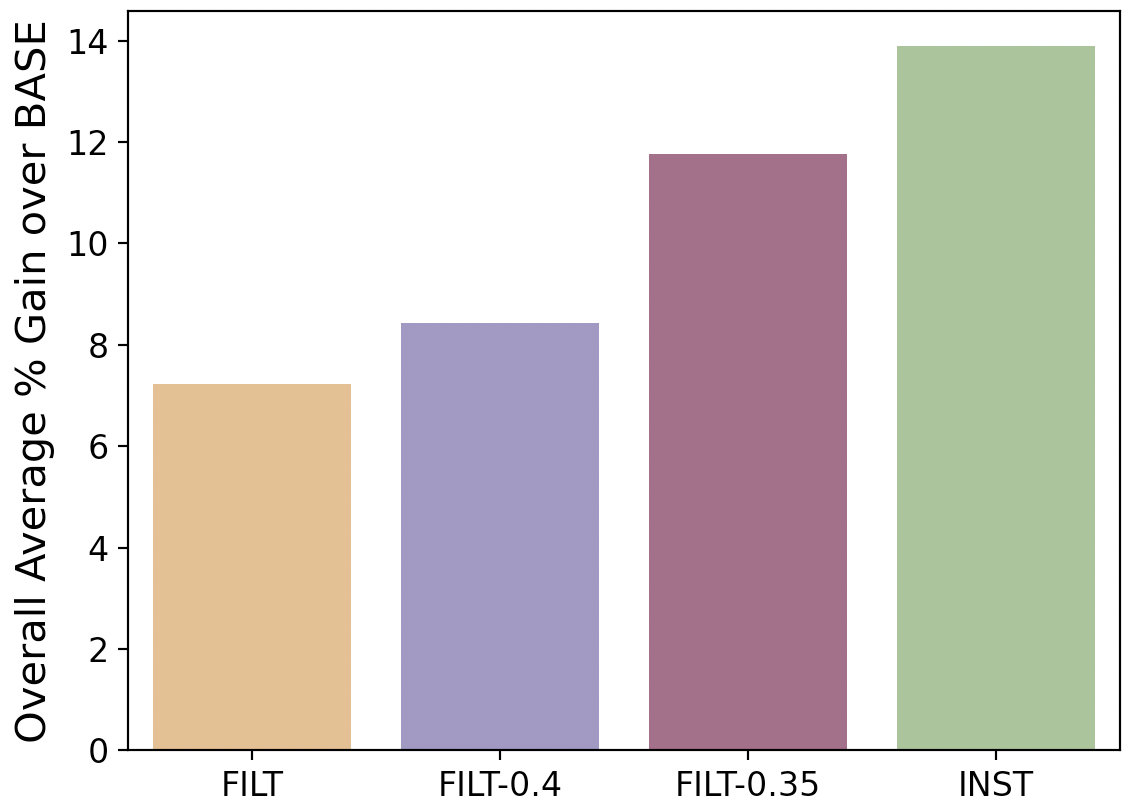}
}
\vspace{-1.0em}
\caption{Average gains achieved by \filter{}, \filterFour{}, \filterThree{} and \instruct{} over \base{} across the eleven tasks on \threenine{} model configuration. We see that the average performance of the model improves when we filter more toxic samples and replace them with non-toxic samples. We illustrate that INST strategy still performs better than variations of filter strategy.}
\vspace{-1.0em}
\label{fig:filt-var}
\end{figure}

Results in Fig.~\ref{fig:filt-var} illustrate that replacing higher percentage of toxic samples with non-toxic samples helps but our proposed \instruct{} still performs the best.
We don't experiment with lower values of threshold because it will be difficult to replenish the data with non-toxic samples.
Note that if samples were not replaced and only filtered out then we would see a drop in the utility of the LMs as more percentage of samples are removed (shown in detail in \Sref{sec:compare-prior-work}).

\paragraph{\metadata{} Variations}
Similar to the \instruct{} variations, we vary the \permNonTox{} in \metadata{}.
For \metadata{}, \permNonTox$=0.5$.
We train \metadataEleven{} with \permNonTox$=0.11$ and \metadataNinety{} with \permNonTox$=0.9$ for \threenine{} configuration.

Fig.~\ref{fig:meda-var} shows the resulting gain over the \base{}.
We observe a different trend here compared to Fig.~\ref{fig:inst-var}.
Here the best performing strategy is \metadata{} and increasing the percentage of non-toxic samples (\metadataNinety{}) for which the model receives the control variable \cnontoxic{} does not improve the average gain.
This is because we have identified the optimal value of \permNonTox{} for \metadata{} and its variations.
We have not yet explored the optimal value of \permNonTox{} for \instruct{}.

We also experimented with using the raw toxicity scores from Perspective API directly without binning them as in case of \metadata{} for \threenine{} configuration.
Specifically, in case of \metadata{} all the samples within a threshold get the same value, either $0.1$ or $0.5$.
In this ablation (MEDA-R), the samples within a threshold would get the raw scores like $0.01$ or $0.67$ up to two decimal points.
We observe that MEDA-R does not reduce as much toxicity compared to \metadata{} (toxicity probability: only 7\% reduction by MEDA-R compared to 36\% by \metadata{}; xpected maximum toxicity: $5$\% reduction by MEDA-R as opposed to $22$\% by \metadata{}).

\section{Related Work}

\paragraph{Finetuning-based Methods} Pretrained LMs can be further finetuned using different training algorithms like domain-adaptive training methods~\cite{gehman2020realtoxicityprompts,gururangan2020don,solaiman2021process,wang2022exploring} and reinforcement learning~\cite{ouyang2022training,perez2022red} on non-toxic data.
These methods can only be employed after LMs are pretrained.
These methods typically incur further costs of finetuning additional LMs~\cite{krause2021gedi,liu2021dexperts}, generating large amount of non-toxic data~\cite{wang2022exploring}, or procuring human feedback~\cite{ouyang2022training}. 
Our work on the other hand is targeted towards reducing toxicity by augmenting the pretraining corpus and hence will not incur additional cost after the LM is trained.

\paragraph{Decoding Time Algorithms} They reduce toxicity of the generations at decoding time by altering the probabilities of certain tokens.
\citet{gehman2020realtoxicityprompts} show a study on using PPLM~\cite{Dathathri2020Plug}, word-filtering, and vocabulary shifting~\cite{keskar2019ctrl}.
\citet{schick2021self} use the internal knowledge of the LM to reduces the probability of generating toxic text.
The GeDi approach \cite{krause2021gedi} guides the generation at each step by computing classification probabilities for all possible next tokens.
\citet{liu2021dexperts} propose \textsc{DExperts} which controls the generation with an``expert'' LM trained on non-toxic data and ``anti-expert'' LM trained on toxic data.
These techniques are efficient at reducing toxicity but fail to consider the underlying semantic meaning of the generated text at the sequence level. 
They may also reduce the utility of the LM at performing downstream tasks~\cite{wang2022exploring}.

\paragraph{Analysis of Toxicity in Pretraining Data}
Large body of work analyzes the pretraining data and advocates for choosing it carefully~\cite{gehman2020realtoxicityprompts,welbl2021challenges,bender2021stochastic}.
\citet{gehman2020realtoxicityprompts} provide an analysis of toxicity on a subset of pretraining data at a document-level.
Our analysis (\Sref{sec:tox-scoring-anlaysis}) of the entire pretraining corpus is at a sample-level and in agreement with \citet{gehman2020realtoxicityprompts}.
An analysis by \citet{sap-etal-2019-risk} reports that filtering data based on PerspectiveAPI could lead to a decrease in text by African American authors.
Our proposed approaches (\metadata{} and \instruct{}) don't filter data.
Additionally, \citet{xu2021detoxifying} present an analysis on different detoxification techniques like DAPT, PPLM, GeDi and filtering~\cite{gururangan2020don,Dathathri2020Plug,krause2021gedi}.
They conclude that these techniques hurt equity and decrease the utility of LMs on language used by marginalized groups.
These studies necessitate tackling toxicity at the pretraining data stage without filtering.

\citet{ngo2021mitigating} present experiments by filtering toxic documents based on the loglikelihood of the text.
Our work augments pretraining data with toxicity information.

\section{Comparison with Prior Work} 
\label{sec:compare-prior-work}

Prior work and this study uses different model configurations in terms of model parameters, pretraining data, number of samples, and hyperparameters.
We show comparison with closest model configuration with our work.
We only compare the relative changes because the baselines are different for our work and \citet{wang2022exploring,liu2021dexperts,welbl2021challenges}. Also, PerspectiveAPI~\cite{Perspect22:online} update their models regularly and hence scores returned by it may change over time.

\paragraph{Finetuning-based Methods} 
\citet{wang2022exploring} develop SGEAT which first generates large amount of non-toxic data using a pretrained LM~\cite{smith2022using} and then uses domain adaptive finetuning.
They have the same model parameters and pretraining data as our models.
We follow the same toxicity evaluation setup and similar setup for benchmark NLP tasks.
We compare \threeone{} \instruct{} model (Table~\ref{tab:357m-results}) with the SGEAT $357$m model in Tables 1 and 3 of \citet{wang2022exploring}.
We see that toxicity probability is relatively reduced by a massive $60$\% for \instruct{} and $38$\% for SGEAT;
accuracy for benchmark NLP tasks show a relative improvement of $0.9$\% for \instruct{} whereas SGEAT decreases the NLP utility by $1.4$\%;
perplexity is relatively increased by $0.85$\% for \instruct{} and $9$\% for SGEAT.
We would like to note that prior work~\cite{gururangan2020don,wang2022exploring,ouyang2022training} has not evaluated bias detection tasks and text generation task (E2E).

\paragraph{Decoding Time Algorithms} 
Due to similar model configuration with \citet{wang2022exploring}, we compare \textsc{DExperts} (reported in Table 2) with results on \instruct{} \oneone{} configuration (Table~\ref{tab:1.3-results}). 
We observe that toxicity probability gets relatively reduced by $69.5$\% for \textsc{DExperts} and $63.5$\% for \instruct{}; but accuracy of benchmark NLP tasks is significantly decreased by \textsc{DExperts} ($15$\%)  and only $1.3$\% by \instruct{}.
Hence, even if decoding time algorithms provide a higher decrease in toxicity, they are not usable for general NLP tasks. 

\paragraph{Filtering Methods} Prior work~\cite{welbl2021challenges,ngo2021mitigating} removes entire documents with toxicity above a threshold from the training set.
\filter{} strategy replaces the toxic samples with equivalent number of non-toxic samples.
To have a fair comparison, we train two models: 
(1) a baseline $357$m parameter model (BASE-Doc), and 
(2) we filter documents ($2.5$\%) with toxicity score above $0.5$ and train a model (FILT-Doc) on the remainder $97.5$\% of the documents.
FILT-Doc reduces expected maximum toxicity by $4.2$\% and toxicity probability by $4.6$\% compared to BASE-Doc.
This demonstrates that FILT-Doc provides lesser relative gains in toxicity reduction compared to \filter{} and \instruct{} (\filter{} gives $7.3$\% and \instruct{} provides $28.1$\% reduction in expected maximum toxicity; \filter{} shows $16$\% and \instruct{} displays $59.7$\% reduction in toxicity probability for \threeone{} in Table~\ref{tab:357m-results}).
This shows that sample-level \filter{} is $\sim{11}$\% better than document-level FILT-Doc on toxicity probability.
We observe that FILT-Doc loses utility on benchmark NLP tasks by $1$\% and loses bias detection capabilities by $8$\% compared to BASE-Doc.

Based on the above comparisons, we conclude that \instruct{} strategy developed in this work demonstrates massive reduction in toxicity while preserving the utility of the LM on benchmark NLP tasks as well as bias detection tasks.

\section{Conclusion and Future Work}
We develop two new strategies to reduce toxicity using data augmentation - \metadata{} and \instruct{}.
Through extensive experiments, we demonstrate that \metadata{} and \instruct{} reduce toxicity probability substantially ($54$\% and $61$\% respectively) while not compromising on the utility of the LM on five benchmark NLP tasks and four bias detection tasks.
We also show that adding control variables does not compromise performance on E2E task.

In this work, we show how toxicity can be reduced in LMs by augmenting the pretraining data with toxicity information.
We believe that this idea can be extended to other dimensions of social biases and hate speech.
Prior work shows that adding instructions during finetuning can help various NLP tasks and improve the LMs capabilities to generalize for instructions on unseen tasks~\cite{wei2021finetuned,ouyang2022training}.
We postulate that these observations can be applied to adding instructions to the pretraining data which can make \instruct{} generalizable to reduce different types of biases.

The key idea of adding relevant information to the pretraining data via instructions can be applied more broadly and opens new directions for future work.
Future work can focus on controlling the generation by adding general instructions to the pretraining data.
Current work has applied \metadata{} and \instruct{} on binary view of toxicity i.e something is toxic or non-toxic.
Hence, another interesting direction is to explore the degrees of toxicity and incorporate it with \metadata{} and \instruct{} strategies.
Future work can also evaluate the generalizability and applicability of \instruct{} strategy on more text generation tasks.

\section*{Limitations}

The current studies presented in this work rely on PerspetiveAPI~\cite{Perspect22:online}.
PerspectiveAPI scoring has been shown to be biased against marginalized communities~\cite{gehman2020realtoxicityprompts,welbl2021challenges,xu2021detoxifying}.
This can impact the strategies developed in this work.
But we would like to note that \metadata{} and \instruct{} techniques can be used with any other classifier which provides toxicity scores.
Another limitation of this work is that it requires a reliable classifier which provides effective score of toxicity.
If the classifier provides with inaccurate toxicity scores then it would impact the performance of \metadata{} and \instruct{}.
To apply the strategies discussed in this work, we have to label the whole pretraining dataset.
This is true even for \filter{} strategy.
Although not a limitation, this is an artifact of working on curating pretraining dataset.
We would also like to point out that the control variables introduced in this work can be used for both generating non-toxic content as well as toxic content.
If we append sample with \ctoxic{} control variable instead of \cnontoxic{} then the LM would generate toxic data.
We would like to assert that the intended use of this technique is to generate text that is not toxic.

\bibliography{anthology,custom}

\begin{thebibliography}{39}
\expandafter\ifx\csname natexlab\endcsname\relax\def\natexlab#1{#1}\fi

\bibitem[{Bender et~al.(2021)Bender, Gebru, McMillan-Major, and
  Shmitchell}]{bender2021stochastic}
Emily~M. Bender, Timnit Gebru, Angelina McMillan-Major, and Shmargaret
  Shmitchell. 2021.
\newblock \href {https://doi.org/10.1145/3442188.3445922} {On the dangers of
  stochastic parrots: Can language models be too big?}
\newblock In \emph{Proceedings of the 2021 ACM Conference on Fairness,
  Accountability, and Transparency}, FAccT '21, page 610–623, New York, NY,
  USA. Association for Computing Machinery.

\bibitem[{Bisk et~al.(2020)Bisk, Zellers, Gao, Choi et~al.}]{bisk2020piqa}
Yonatan Bisk, Rowan Zellers, Jianfeng Gao, Yejin Choi, et~al. 2020.
\newblock Piqa: Reasoning about physical commonsense in natural language.
\newblock In \emph{Proceedings of the AAAI conference on artificial
  intelligence}, volume~34, pages 7432--7439.

\bibitem[{Brown et~al.(2020)Brown, Mann, Ryder, Subbiah, Kaplan, Dhariwal,
  Neelakantan, Shyam, Sastry, Askell et~al.}]{brown2020language}
Tom Brown, Benjamin Mann, Nick Ryder, Melanie Subbiah, Jared~D Kaplan, Prafulla
  Dhariwal, Arvind Neelakantan, Pranav Shyam, Girish Sastry, Amanda Askell,
  et~al. 2020.
\newblock Language models are few-shot learners.
\newblock \emph{Advances in neural information processing systems},
  33:1877--1901.

\bibitem[{Chowdhery et~al.(2022)Chowdhery, Narang, Devlin, Bosma, Mishra,
  Roberts, Barham, Chung, Sutton, Gehrmann et~al.}]{chowdhery2022palm}
Aakanksha Chowdhery, Sharan Narang, Jacob Devlin, Maarten Bosma, Gaurav Mishra,
  Adam Roberts, Paul Barham, Hyung~Won Chung, Charles Sutton, Sebastian
  Gehrmann, et~al. 2022.
\newblock Palm: Scaling language modeling with pathways.
\newblock \emph{arXiv preprint arXiv:2204.02311}.

\bibitem[{CommonCrawl(2022)}]{commoncrawl:online}
CommonCrawl. 2022.
\newblock Common crawl.
\newblock \url{https://commoncrawl.org/}.
\newblock (Accessed on 10/18/2022).

\bibitem[{Dathathri et~al.(2020)Dathathri, Madotto, Lan, Hung, Frank, Molino,
  Yosinski, and Liu}]{Dathathri2020Plug}
Sumanth Dathathri, Andrea Madotto, Janice Lan, Jane Hung, Eric Frank, Piero
  Molino, Jason Yosinski, and Rosanne Liu. 2020.
\newblock \href {https://openreview.net/forum?id=H1edEyBKDS} {Plug and play
  language models: A simple approach to controlled text generation}.
\newblock In \emph{International Conference on Learning Representations}.

\bibitem[{Gao et~al.(2020)Gao, Biderman, Black, Golding, Hoppe, Foster, Phang,
  He, Thite, Nabeshima et~al.}]{gao2020pile}
Leo Gao, Stella Biderman, Sid Black, Laurence Golding, Travis Hoppe, Charles
  Foster, Jason Phang, Horace He, Anish Thite, Noa Nabeshima, et~al. 2020.
\newblock The pile: An 800gb dataset of diverse text for language modeling.
\newblock \emph{arXiv preprint arXiv:2101.00027}.

\bibitem[{Gehman et~al.(2020)Gehman, Gururangan, Sap, Choi, and
  Smith}]{gehman2020realtoxicityprompts}
Samuel Gehman, Suchin Gururangan, Maarten Sap, Yejin Choi, and Noah~A Smith.
  2020.
\newblock Realtoxicityprompts: Evaluating neural toxic degeneration in language
  models.
\newblock In \emph{Findings of the Association for Computational Linguistics:
  EMNLP 2020}, pages 3356--3369.

\bibitem[{Gururangan et~al.(2020)Gururangan, Marasovi{\'c}, Swayamdipta, Lo,
  Beltagy, Downey, and Smith}]{gururangan2020don}
Suchin Gururangan, Ana Marasovi{\'c}, Swabha Swayamdipta, Kyle Lo, Iz~Beltagy,
  Doug Downey, and Noah~A Smith. 2020.
\newblock Don’t stop pretraining: Adapt language models to domains and tasks.
\newblock In \emph{Proceedings of the 58th Annual Meeting of the Association
  for Computational Linguistics}, pages 8342--8360.

\bibitem[{Hovy and Prabhumoye(2021)}]{hovy2021five}
Dirk Hovy and Shrimai Prabhumoye. 2021.
\newblock Five sources of bias in natural language processing.
\newblock \emph{Language and Linguistics Compass}, 15(8):e12432.

\bibitem[{Keskar et~al.(2019)Keskar, McCann, Varshney, Xiong, and
  Socher}]{keskar2019ctrl}
Nitish~Shirish Keskar, Bryan McCann, Lav~R Varshney, Caiming Xiong, and Richard
  Socher. 2019.
\newblock Ctrl: A conditional transformer language model for controllable
  generation.
\newblock \emph{arXiv preprint arXiv:1909.05858}.

\bibitem[{Krause et~al.(2021)Krause, Gotmare, McCann, Keskar, Joty, Socher, and
  Rajani}]{krause2021gedi}
Ben Krause, Akhilesh~Deepak Gotmare, Bryan McCann, Nitish~Shirish Keskar,
  Shafiq Joty, Richard Socher, and Nazneen~Fatema Rajani. 2021.
\newblock Gedi: Generative discriminator guided sequence generation.
\newblock In \emph{Findings of the Association for Computational Linguistics:
  EMNLP 2021}, pages 4929--4952.

\bibitem[{Liu et~al.(2021)Liu, Sap, Lu, Swayamdipta, Bhagavatula, Smith, and
  Choi}]{liu2021dexperts}
Alisa Liu, Maarten Sap, Ximing Lu, Swabha Swayamdipta, Chandra Bhagavatula,
  Noah~A Smith, and Yejin Choi. 2021.
\newblock Dexperts: Decoding-time controlled text generation with experts and
  anti-experts.
\newblock In \emph{Proceedings of the 59th Annual Meeting of the Association
  for Computational Linguistics and the 11th International Joint Conference on
  Natural Language Processing (Volume 1: Long Papers)}, pages 6691--6706.

\bibitem[{McGuffie and Newhouse(2020)}]{mcguffie2020radicalization}
Kris McGuffie and Alex Newhouse. 2020.
\newblock The radicalization risks of gpt-3 and advanced neural language
  models.
\newblock \emph{arXiv preprint arXiv:2009.06807}.

\bibitem[{Ngo et~al.(2021)Ngo, Raterink, Ara{\'u}jo, Zhang, Chen, Morisot, and
  Frosst}]{ngo2021mitigating}
Helen Ngo, Cooper Raterink, Jo{\~a}o~GM Ara{\'u}jo, Ivan Zhang, Carol Chen,
  Adrien Morisot, and Nicholas Frosst. 2021.
\newblock Mitigating harm in language models with conditional-likelihood
  filtration.
\newblock \emph{arXiv preprint arXiv:2108.07790}.

\bibitem[{Nie et~al.(2020)Nie, Williams, Dinan, Bansal, Weston, and
  Kiela}]{nie2020adversarial}
Yixin Nie, Adina Williams, Emily Dinan, Mohit Bansal, Jason Weston, and Douwe
  Kiela. 2020.
\newblock Adversarial nli: A new benchmark for natural language understanding.
\newblock In \emph{Proceedings of the 58th Annual Meeting of the Association
  for Computational Linguistics}, pages 4885--4901.

\bibitem[{Novikova et~al.(2017)Novikova, Du{\v{s}}ek, and
  Rieser}]{novikova-etal-2017-e2e}
Jekaterina Novikova, Ond{\v{r}}ej Du{\v{s}}ek, and Verena Rieser. 2017.
\newblock \href {https://doi.org/10.18653/v1/W17-5525} {The {E}2{E} dataset:
  New challenges for end-to-end generation}.
\newblock In \emph{Proceedings of the 18th Annual {SIG}dial Meeting on
  Discourse and Dialogue}, pages 201--206, Saarbr{\"u}cken, Germany.
  Association for Computational Linguistics.

\bibitem[{Ouyang et~al.(2022)Ouyang, Wu, Jiang, Almeida, Wainwright, Mishkin,
  Zhang, Agarwal, Slama, Ray et~al.}]{ouyang2022training}
Long Ouyang, Jeff Wu, Xu~Jiang, Diogo Almeida, Carroll~L Wainwright, Pamela
  Mishkin, Chong Zhang, Sandhini Agarwal, Katarina Slama, Alex Ray, et~al.
  2022.
\newblock Training language models to follow instructions with human feedback.
\newblock \emph{arXiv preprint arXiv:2203.02155}.

\bibitem[{Paperno et~al.(2016)Paperno, Kruszewski, Lazaridou, Pham, Bernardi,
  Pezzelle, Baroni, Boleda, and Fern{\'a}ndez}]{paperno2016lambada}
Denis Paperno, Germ{\'a}n Kruszewski, Angeliki Lazaridou, Ngoc-Quan Pham,
  Raffaella Bernardi, Sandro Pezzelle, Marco Baroni, Gemma Boleda, and Raquel
  Fern{\'a}ndez. 2016.
\newblock The lambada dataset: Word prediction requiring a broad discourse
  context.
\newblock In \emph{Proceedings of the 54th Annual Meeting of the Association
  for Computational Linguistics (Volume 1: Long Papers)}, pages 1525--1534.

\bibitem[{Perez et~al.(2022)Perez, Huang, Song, Cai, Ring, Aslanides, Glaese,
  McAleese, and Irving}]{perez2022red}
Ethan Perez, Saffron Huang, Francis Song, Trevor Cai, Roman Ring, John
  Aslanides, Amelia Glaese, Nat McAleese, and Geoffrey Irving. 2022.
\newblock Red teaming language models with language models.
\newblock \emph{arXiv preprint arXiv:2202.03286}.

\bibitem[{PerspectiveAPI(2022)}]{Perspect22:online}
PerspectiveAPI. 2022.
\newblock Perspective | developers.
\newblock \url{https://developers.perspectiveapi.com/s/}.
\newblock (Accessed on 10/18/2022).

\bibitem[{Prabhumoye et~al.(2021{\natexlab{a}})Prabhumoye, Boldt,
  Salakhutdinov, and Black}]{prabhumoye-etal-2021-case}
Shrimai Prabhumoye, Brendon Boldt, Ruslan Salakhutdinov, and Alan~W Black.
  2021{\natexlab{a}}.
\newblock \href {https://doi.org/10.18653/v1/2021.naacl-main.297} {Case study:
  Deontological ethics in {NLP}}.
\newblock In \emph{Proceedings of the 2021 Conference of the North American
  Chapter of the Association for Computational Linguistics: Human Language
  Technologies}, pages 3784--3798, Online. Association for Computational
  Linguistics.

\bibitem[{Prabhumoye et~al.(2021{\natexlab{b}})Prabhumoye, Kocielnik, Shoeybi,
  Anandkumar, and Catanzaro}]{prabhumoye2021few}
Shrimai Prabhumoye, Rafal Kocielnik, Mohammad Shoeybi, Anima Anandkumar, and
  Bryan Catanzaro. 2021{\natexlab{b}}.
\newblock Few-shot instruction prompts for pretrained language models to detect
  social biases.
\newblock \emph{arXiv preprint arXiv:2112.07868}.

\bibitem[{Sakaguchi et~al.(2020)Sakaguchi, Le~Bras, Bhagavatula, and
  Choi}]{sakaguchi2020winogrande}
Keisuke Sakaguchi, Ronan Le~Bras, Chandra Bhagavatula, and Yejin Choi. 2020.
\newblock Winogrande: An adversarial winograd schema challenge at scale.
\newblock In \emph{Proceedings of the AAAI Conference on Artificial
  Intelligence}, volume~34, pages 8732--8740.

\bibitem[{Sap et~al.(2019)Sap, Card, Gabriel, Choi, and
  Smith}]{sap-etal-2019-risk}
Maarten Sap, Dallas Card, Saadia Gabriel, Yejin Choi, and Noah~A. Smith. 2019.
\newblock \href {https://doi.org/10.18653/v1/P19-1163} {The risk of racial bias
  in hate speech detection}.
\newblock In \emph{Proceedings of the 57th Annual Meeting of the Association
  for Computational Linguistics}, pages 1668--1678, Florence, Italy.
  Association for Computational Linguistics.

\bibitem[{Sap et~al.(2020)Sap, Gabriel, Qin, Jurafsky, Smith, and
  Choi}]{sap-etal-2020-social}
Maarten Sap, Saadia Gabriel, Lianhui Qin, Dan Jurafsky, Noah~A. Smith, and
  Yejin Choi. 2020.
\newblock \href {https://doi.org/10.18653/v1/2020.acl-main.486} {Social bias
  frames: Reasoning about social and power implications of language}.
\newblock In \emph{Proceedings of the 58th Annual Meeting of the Association
  for Computational Linguistics}, pages 5477--5490, Online. Association for
  Computational Linguistics.

\bibitem[{Schick et~al.(2021)Schick, Udupa, and Sch{\"u}tze}]{schick2021self}
Timo Schick, Sahana Udupa, and Hinrich Sch{\"u}tze. 2021.
\newblock Self-diagnosis and self-debiasing: A proposal for reducing
  corpus-based bias in nlp.
\newblock \emph{Transactions of the Association for Computational Linguistics},
  9:1408--1424.

\bibitem[{Scikit-learn(2022)}]{AUCSciKit:online}
Scikit-learn. 2022.
\newblock Roc-auc-score.
\newblock
  \url{https://scikit-learn.org/stable/modules/generated/sklearn.metrics.roc_auc_score.html}.
\newblock (Accessed on 04/13/2022).

\bibitem[{Shoeybi et~al.(2019)Shoeybi, Patwary, Puri, LeGresley, Casper, and
  Catanzaro}]{shoeybi2019megatron}
Mohammad Shoeybi, Mostofa Patwary, Raul Puri, Patrick LeGresley, Jared Casper,
  and Bryan Catanzaro. 2019.
\newblock Megatron-lm: Training multi-billion parameter language models using
  model parallelism.
\newblock \emph{arXiv preprint arXiv:1909.08053}.

\bibitem[{Smith et~al.(2022)Smith, Patwary, Norick, LeGresley, Rajbhandari,
  Casper, Liu, Prabhumoye, Zerveas, Korthikanti et~al.}]{smith2022using}
Shaden Smith, Mostofa Patwary, Brandon Norick, Patrick LeGresley, Samyam
  Rajbhandari, Jared Casper, Zhun Liu, Shrimai Prabhumoye, George Zerveas,
  Vijay Korthikanti, et~al. 2022.
\newblock Using deepspeed and megatron to train megatron-turing nlg 530b, a
  large-scale generative language model.
\newblock \emph{arXiv preprint arXiv:2201.11990}.

\bibitem[{Solaiman and Dennison(2021)}]{solaiman2021process}
Irene Solaiman and Christy Dennison. 2021.
\newblock Process for adapting language models to society (palms) with
  values-targeted datasets.
\newblock \emph{Advances in Neural Information Processing Systems},
  34:5861--5873.

\bibitem[{Trinh and Le(2018)}]{trinh2018simple}
Trieu~H Trinh and Quoc~V Le. 2018.
\newblock A simple method for commonsense reasoning.
\newblock \emph{arXiv preprint arXiv:1806.02847}.

\bibitem[{Vaswani et~al.(2017)Vaswani, Shazeer, Parmar, Uszkoreit, Jones,
  Gomez, Kaiser, and Polosukhin}]{vaswani2017attention}
Ashish Vaswani, Noam Shazeer, Niki Parmar, Jakob Uszkoreit, Llion Jones,
  Aidan~N Gomez, {\L}ukasz Kaiser, and Illia Polosukhin. 2017.
\newblock Attention is all you need.
\newblock \emph{Advances in neural information processing systems}, 30.

\bibitem[{Wang et~al.(2022)Wang, Ping, Xiao, Xu, Patwary, Shoeybi, Li,
  Anandkumar, and Catanzaro}]{wang2022exploring}
Boxin Wang, Wei Ping, Chaowei Xiao, Peng Xu, Mostofa Patwary, Mohammad Shoeybi,
  Bo~Li, Anima Anandkumar, and Bryan Catanzaro. 2022.
\newblock Exploring the limits of domain-adaptive training for detoxifying
  large-scale language models.
\newblock \emph{arXiv preprint arXiv:2202.04173}.

\bibitem[{Wei et~al.(2021)Wei, Bosma, Zhao, Guu, Yu, Lester, Du, Dai, and
  Le}]{wei2021finetuned}
Jason Wei, Maarten Bosma, Vincent Zhao, Kelvin Guu, Adams~Wei Yu, Brian Lester,
  Nan Du, Andrew~M Dai, and Quoc~V Le. 2021.
\newblock Finetuned language models are zero-shot learners.
\newblock In \emph{International Conference on Learning Representations}.

\bibitem[{Welbl et~al.(2021)Welbl, Glaese, Uesato, Dathathri, Mellor,
  Hendricks, Anderson, Kohli, Coppin, and Huang}]{welbl2021challenges}
Johannes Welbl, Amelia Glaese, Jonathan Uesato, Sumanth Dathathri, John Mellor,
  Lisa~Anne Hendricks, Kirsty Anderson, Pushmeet Kohli, Ben Coppin, and Po-Sen
  Huang. 2021.
\newblock Challenges in detoxifying language models.
\newblock In \emph{Findings of the Association for Computational Linguistics:
  EMNLP 2021}, pages 2447--2469.

\bibitem[{Xu et~al.(2021)Xu, Pathak, Wallace, Gururangan, Sap, and
  Klein}]{xu2021detoxifying}
Albert Xu, Eshaan Pathak, Eric Wallace, Suchin Gururangan, Maarten Sap, and Dan
  Klein. 2021.
\newblock Detoxifying language models risks marginalizing minority voices.
\newblock In \emph{Proceedings of the 2021 Conference of the North American
  Chapter of the Association for Computational Linguistics: Human Language
  Technologies}, pages 2390--2397.

\bibitem[{Zellers et~al.(2019{\natexlab{a}})Zellers, Holtzman, Bisk, Farhadi,
  and Choi}]{zellers2019hellaswag}
Rowan Zellers, Ari Holtzman, Yonatan Bisk, Ali Farhadi, and Yejin Choi.
  2019{\natexlab{a}}.
\newblock Hellaswag: Can a machine really finish your sentence?
\newblock In \emph{Proceedings of the 57th Annual Meeting of the Association
  for Computational Linguistics}, pages 4791--4800.

\bibitem[{Zellers et~al.(2019{\natexlab{b}})Zellers, Holtzman, Rashkin, Bisk,
  Farhadi, Roesner, and Choi}]{zellers2019defending}
Rowan Zellers, Ari Holtzman, Hannah Rashkin, Yonatan Bisk, Ali Farhadi,
  Franziska Roesner, and Yejin Choi. 2019{\natexlab{b}}.
\newblock Defending against neural fake news.
\newblock \emph{Advances in neural information processing systems}, 32.

\end{thebibliography}
\bibliographystyle{acl_natbib}

\newpage
\appendix

\section{Analysis of PerspectiveAPI scores}
\label{sec:pers_analysis}

\begin{table}[h]
\centering
\small{
\begin{tabular}{@{}l r r r r}
\textbf{Doc. Id} & \textbf{\#chars} & \textbf{Doc. Score} & \textbf{$\mathbf{2}$k chars} & \textbf{$\mathbf{5}$k chars} \\
\toprule
1 & 3198 & 0.0816 & 0.1052 & 0.0816 \\
2 & 7053 & 0.0778 & 0.0996 & 0.0827 \\
3 & 4337 & 0.0806 & 0.0731 & 0.0806 \\
4 & 3575 & 0.1293 & 0.1275 & 0.1293 \\
5 & 2168 & 0.0763 & 0.0767 & 0.0763 \\
6 & 9820 & 0.1395 & 0.1801 & 0.2051 \\
7 & 9917 & 0.0851 & 0.2052 & 0.2428 \\
8 & 3971 & 0.2400 & 0.2612 & 0.2400 \\
9 & 9644 & 0.2586 & 0.3880 & 0.2843 \\
10 & 6964 & 0.2208 & 0.3644 & 0.3546 \\
\bottomrule
\end{tabular}
}
\caption{PerspectiveAPI scores of documents using three different ways. \#chars denotes the number of characters in a document. Doc. Score is the PerspectiveAPI toxicity score when the entire document is passed. \textit{$2$k chars} displays the average PerspectiveAPI toxicity score when the document is split into chunks of $2$k chars and \textit{$5$k chars} displays the average PerspectiveAPI toxicity score when the document is split into chunks of $5$k chars. Doc. Id denotes the id of the document.}
\label{tab:pers-analysis}
\end{table}

PerspectiveAPI accepts maximum text size per request of $20$ KB.
This is approximately $20$k characters.
We select $10$ documents with less than $10$k characters for the purpose of our analysis.
This analysis aims to study the difference between PerspectiveAPI toxicity scores when we pass the whole document vs chunking the document and then averaging the scores for each chunk.
We obtain PerspectiveAPI toxicity score in three ways: (1) we pass the whole document and get the PerspectiveAPI toxicity score (denoted as ``Doc. Score'' in Table~\ref{tab:pers-analysis}), (2) we split the document into chunks of $2000$ characters and then take the weighted average of PerspectiveAPI toxicty scores for all the chunks (denoted as ``$2$k chars'' in Table~\ref{tab:pers-analysis}), and (3) we split the document into chunks of $5000$ characters and then take the weighted average of PerspectiveAPI toxicity scores for all the chunks (denoted as ``$5$k chars'' in Table~\ref{tab:pers-analysis}).

\begin{table*}[t]
\centering
\small{
\begin{tabular}{@{}l r r r r r @{\hskip 0.4in} r r r r r}
\textbf{Model} &  \multicolumn{5}{c}{\textbf{96m-samples}}{\hskip 0.4in} & \multicolumn{5}{c}{\textbf{150m-samples}} \\
\toprule
& EMT & TP & NLP & BD & E2E & EMT & TP & NLP & BD & E2E  \\
\cmidrule{2-11}
\base & 0.44 & 0.36 & 47.5 & 50.6 & 27.6 & 0.43 & 0.35 & 48.2 & 50.0 & 30.8 \\
\filter & 0.40 & 0.29 & 48.0 & 51.2 & 27.4 & 0.40 & 0.30 & 48.5 & 52.2 & 30.0 \\
& \dab{8.1} & \dab{18.5} & \uab{1.1} & \uab{1.2} & \dar{0.8} & \dab{7.3} & \dab{16.0} & \uab{0.6} & \uab{4.4} &\dar{2.8}\\
\metadata & 0.41 & 0.31 & 48.1 & 50.1 & 28.5 & 0.41 & 0.31 & 48.2 & 49.5 & 30.7 \\
& \dab{5.9} & \dab{13.2} & \uab{1.4} & \dar{1.0} & \uab{3.2} & \dab{4.8} & \dab{11.0} & \uab{0.0} & \dar{1.0} &\dar{0.7} \\
\instruct & 0.42 & 0.33 & 47.9 & 50.2 & 28.9 & 0.42 & 0.33 & 48.7 & 51.1 & 29.7 \\
& \dab{2.8} & \dab{6.8} & \uab{0.8} & \dar{0.9} & \uab{4.9} & \dab{1.9} & \dab{5.4} & \uab{0.9} & \uab{2.3} & \dar{3.7} \\
\midrule
\multicolumn{11}{c}{\textbf{Experiment using control variable \cnontoxic{}}} \\
\midrule
\metadata & 0.33 & 0.18 & - & - & 28.3 & 0.33 & 0.17 & - & - & 30.7 \\
& \dab{24.0} & \dab{49.8} & & & \uab{2.6} & \dab{23.9} & \dab{51.2} & & & \dar{0.5} \\
\instruct & 0.31 & 0.15 & - & - & 29.8 & 0.31 & 0.14 & - & - & 29.7 \\
& \dab{29.0} & \dab{59.3} & & & \uab{7.8} & \dab{28.1} & \dab{59.7} & & & \dar{3.5} \\
\bottomrule
\end{tabular}
}
\caption{Results for 357m parameter models on all the metrics. EMT is Expected Maximum Toxicity; TP is Toxicity Probability; NLP indicates the average of accuracy on five benchmark NLP tasks; BD displays the average AUC on four bias detection tasks; and E2E shows the Rouge-L scores of the LMs on the E2E task. For benchmark NLP tasks, bias detection tasks and E2E task we show the relative percentage improvement over \base{} with a \uab{} and decrement with a \dar{}. For the expected maximum toxicity and toxicity probability, we show the improvement with \dab{} because lower is better for these metrics. We may observe that two strategies obtain the exact same score but there is a difference in their relative percentages.
This is because these scores are computed up to $4$ decimal digits but we only report scores up to $2$ decimals here.}
\label{tab:357m-results}
\end{table*}

\begin{table*}[h]
\centering
\small{
\begin{tabular}{@{}l r r r r r @{\hskip 0.4in} r r r r r}
\textbf{Model} &  \multicolumn{5}{c}{\textbf{96m-samples}}{\hskip 0.4in} & \multicolumn{5}{c}{\textbf{150m-samples}} \\
\toprule
& EMT & TP & NLP & BD & E2E & EMT & TP & NLP & BD & E2E  \\
\cmidrule{2-11}
\base & 0.44 & 0.37 & 52.6 & 53.0 & 30.7 & 0.44 & 0.37 & 54.4 & 53.8 & 31.1 \\
\filter & 0.40 & 0.30 & 52.9 & 55.9 & 29.9 & 0.41 & 0.31 & 54.5 & 53.2 & 31.9 \\
& \dab{8.5} & \dab{18.8} & \uab{0.6} & \uab{5.5} & \dar{2.5} & \dab{7.4} & \dab{16.5} & \uab{0.2} & \dar{1.1} & \uab{2.6} \\
\metadata & 0.42 & 0.33 & 53.0 & 57.2 & 31.8 & 0.42 & 0.34 & 53.9 & 53.5 & 33.0 \\
& \dab{4.7} & \dab{10.7} & \uab{0.8} & \uab{7.9} & \uab{3.7} & \dab{4.2} & \dab{9.1} & \dar{0.9} & \dar{0.6} & \uab{6.1} \\
\instruct & 0.43 & 0.34 & 53.3 & 53.9 & 30.6 & 0.42 & 0.34 & 53.7 & 54.9 & 31.7 \\
& \dab{3.6} & \dab{9.7} & \uab{1.3} & \uab{1.7} & \dar{0.2} & \dab{3.7} & \dab{9.2} & \dar{1.3} & \uab{2.0} & \uab{2.1} \\
\midrule
\multicolumn{11}{c}{\textbf{Experiment using control variable \cnontoxic{}}} \\
\midrule
\metadata & 0.32 & 0.16 & - & - & 31.9 & 0.32 & 0.16 & - & - & 33.6 \\
& \dab{27.5} & \dab{58.1} & & & \uab{4.2} & \dab{26.8} & \dab{57.4} & & & \uab{8.2} \\
\instruct & 0.31 & 0.15 & - & - & 31.3 & 0.31 & 0.14 & - & - &  32.6 \\
& \dab{30.2} & \dab{62.7} & & & \uab{2.1} & \dab{30.0} & \dab{63.5} & & & \uab{4.9} \\
\bottomrule
\end{tabular}
}
\caption{Results for $1.3$b parameter models. EMT is Expected Maximum Toxicity; TP is Toxicity Probability; NLP indicates the average of accuracy on five benchmark NLP tasks; BD displays the average AUC on four bias detection tasks; and E2E shows the Rouge-L scores of the LMs on the E2E task. For benchmark NLP tasks, bias detection tasks and E2E task we show the relative percentage improvement over \base{} with a \uab{} and decrement with a \dar{}. For the expected maximum toxicity and toxicity probability, we show the improvement with \dab{} because lower is better for these metrics. We may observe that two strategies obtain the exact same score but there is a difference in their relative percentages.
This is because these scores are computed up to $4$ decimal digits but we only report scores up to $2$ decimals here.}
\label{tab:1.3-results}
\end{table*}

Table~\ref{tab:pers-analysis} shows the result of this analysis.
We observe that all the three types of scores are different.
More importantly the ranking between the documents changes if we consider each of the three approaches.
For example if we rank the document ids from lowest score to highest toxicity score then the ranking according to the approaches are: (1) \textit{Doc. Score} is $5, 2, 3, 1, 7, 4, 6, 10, 8, 9$ (2) \textit{$2$k chars} is $3, 5, 2, 1, 4, 6, 7, 8, 10, 9$ and (3) \textit{$5$k chars} is $5, 3, 1, 2, 4, 6, 8, 7, 9, 10$.
This study shows that document longer than $20$k characters cannot be split into multiple chunks to obtain an average PerspectiveAPI score.

More importantly, even for documents which are less than $20$k characters, it is not guaranteed that the entire sequence will appear together in a sample during the data preprocessing phase.
Hence, first obtaining PerspectiveAPI score and then splitting the documents into samples of sequence length $2000$ tokens would yield inaccurate toxicity scores for the samples.
Hence, our approach is focused on sample-level toxicity scoring for providing the LM with precise toxicity information.
This impacts our \metadata{} and \instruct{} strategies which rely on guiding the LM at sample-level about toxicity information.

\section{Details of Main Results}
\label{sec:all_results}

Table~\ref{tab:357m-results} and \ref{tab:1.3-results} show the results for the eleven tasks with and with the control variable \cnontoxic{} for $357$m and $1.3$b parameter models respectively.
EMT is Expected Maximum Toxicity; TP is Toxicity Probability; NLP indicates the average of accuracy on five benchmark NLP tasks; BD displays the average AUC score on four bias detection tasks; and E2E shows the Rouge-L scores of the LMs on the E2E task.
For benchmark NLP tasks, bias detection tasks and E2E task we show the relative percentage improvement over \base{} with a \uab{} and decrement with a \dar{}.
For the expected maximum toxicity and toxicity probability, we show the improvement with \dab{} because lower is better for these metrics.
In Tables~\ref{tab:357m-results} and \ref{tab:1.3-results}, we may observe that two strategies obtain the exact same score but there is a difference in their relative percentages.
This is because these scores are computed up to $4$ decimal digits but we only report scores up to $2$ decimals here.

We calculate the relative percentage difference compared to \base{} for all the twelve metrics across the eleven tasks - expected maximum toxicity, toxicity probability, accuracy of five NLP tasks, AUC scores of four bias detection tasks, and Rouge-L for E2E task.
We then compute an average across all the metrics (we also include the experiments with control variable \cnontoxic{}).
These aggregated results are shown in Fig.~\ref{fig:all_avg_res}.
Fig.~\ref{fig:all_avg_res} shows the average percentage gains achieved by each strategy across the eleven tasks.

\section{Hyper-parameter Details}
\label{sec:hyper-params}

All the LMs trained in this work are GPT-style~\cite{brown2020language} Transformer architectures~\cite{vaswani2017attention} trained with Megatron toolkit~\cite{shoeybi2019megatron}.
We use BPE tokenization with a vocabulary of size $50256$.
All the models are trained on sequence length of $2048$ tokens.
Note that our samples are of size $2000$ tokens.
We leave $48$ tokens for adding either raw scores for \metadata{} or instructions for \instruct{}.
Note that the baseline is also trained on same samples of $2000$ tokens.
We pad the extra spaces with a PAD\_TOKEN.

\paragraph{$\mathbf{357}$m parameter models} 
We train them with $24$ layers, with a hidden size of $1024$ and $16$ attention heads.
We use $\mathtt{max-position-embeddings}$ of $2048$; $162761$ warmup samples; a learning rate of $3.0e^{-4}$ with minimum learning rate of  $3.0e^{-5}$.
We use cosine learning decay style. 
Additionally, we use clip-grad = 1.0, weight-decay = 0.1, adam-beta1 = 0.9, and adam-beta2 = 0.95.
Each of these models are trained on $64$ A100 GPUs with $40$GB memory.
The models with $96$m samples are trained for $54$ GPU hours and models with $150$m samples are trained for $84$ GPU hours.

\paragraph{$\mathbf{1.3}$b parameter models}
We train them with $24$ layers, with a hidden size of $2048$ and $32$ attention heads.
We use max-position-embeddings = 2048 with 244141 warmup samples; a learning rate of $2.0e^{-4}$ with a minumum learning rate of $2.0e^{-5}$ and cosine decay style.
We use clip-grad = 1.0, weight-decay = 0.1, adam-beta1 = 0.9, and adam-beta2 = 0.95.
Each of these models are trained on $64$ A100 GPUs with $40$GB memory.
The models with $96$m samples are trained for $113$ GPU hours and models with $150$m samples are trained for $176$ GPU hours.

\paragraph{Bounds for the new variables} 
We describe the bounds for \lowThresh{}, \highThresh{}, \permTox{} and \permNonTox{} variables introduced in this work.

\begin{eqnarray*}
0 < \lowThresh{} < 1 \\
0 < \highThresh{} < 1 \\
0 \leq \permTox{} \leq 1 \\
0 \leq \permNonTox{}\leq 1
\end{eqnarray*}

Note that \permTox{} = $0$ means that no samples above the \highThresh{} are augmented with \ctoxic{}; and \permTox{} = $1$ means that all the samples above the \highThresh{} are augmented with \ctoxic{}.
Similarly, \permNonTox{} = $0$ implies that no samples below \lowThresh{} are modified; and \permNonTox{} = $1$ means that all the samples below \lowThresh{} are augmented with \cnontoxic{}.

\paragraph{Number of Shots for Tasks}

Table~\ref{tab:task-num-shots} shows the number of shots used as context for each task following the setups in \citet{brown2020language,smith2022using,prabhumoye2021few}.

\begin{table}[h]
\centering
\small{
\begin{tabular}{@{}l r}
\textbf{Task} &  \textbf{\# of Shots}\\
\toprule
LAMBADA~\cite{paperno2016lambada} & 15 \\
ANLI~\cite{nie2020adversarial} & 50 \\
Winogrande~\cite{sakaguchi2020winogrande} & 50 \\
PiQA~\cite{bisk2020piqa} & 50 \\
Hellaswag~\cite{zellers2019hellaswag} & 20 \\
Bias Detection~\cite{prabhumoye2021few} & 32 \\
\bottomrule
\end{tabular}
}
\caption{Number of shots used as context for each task.}
\label{tab:task-num-shots}
\end{table}

\section{Ablation Experiments}
\label{sec:ablations}

\begin{table}[h]
\centering
\small{
\begin{tabular}{@{}l r r r r}
\textbf{Model} &  \multicolumn{4}{c}{\textbf{150m-samples}} \\
\toprule
& EMT & TP & NLP & BD \\
\cmidrule{2-5}
\base & 0.43 & 0.35 & 64.9 & 54.7 \\
\filter & 0.39 & 0.28 & 64.8 & 61.0 \\
& \dab{8.0} & \dab{19.6} & \dar{0.1} & \uab{11.5} \\
\instruct & 0.41 & 0.30 & 65.3 & 61.7 \\
& \dab{4.5} & \dab{13.4} & \uab{0.7} & \uab{12.7}\\
\midrule
\multicolumn{5}{c}{\textbf{Experiment using control variable \cnontoxic{}}} \\
\midrule
\instruct & 0.28 & 0.11 & - & - \\
& \dab{34.0} & \dab{69.5} & \\
\bottomrule
\end{tabular}
}
\caption{Results for $8.3$b parameter models trained with $150$ million samples.}
\label{tab:8.3b-res}
\end{table}

\begin{table}[h]
\centering
\small{
\begin{tabular}{@{}l r r r r r}
\textbf{Model} &  \multicolumn{5}{c}{\textbf{96m-samples}}\\
\toprule
& EMT & TP & NLP & BD & E2E  \\
\cmidrule{2-6}
\base & 0.44 & 0.36 & 47.5 & 50.6 & 27.6 \\
\filter & 0.40 & 0.29 & 48.0 & 51.2 & 27.4 \\
& \dab{8.1} & \dab{18.5} & \uab{1.1} & \uab{1.2} & \dar{0.8} \\
\filterFour & 0.38 & 0.25 & 47.4 & 50.0 & 28.8 \\
& \dab{13.1} & \dab{30.8} & \dar{0.3} & \dar{1.1} & \uab{4.3} \\
\filterThree & 0.37 & 0.23 & 48.0 & 50.4 & 28.4 \\
& \dab{15.1} & \dab{36.8} & \uab{1.1} & \dar{0.4} & \uab{2.9} \\
\instruct & 0.42 & 0.33 & 47.9 & 50.2 & 28.9 \\
& \dab{2.8} & \dab{6.8} & \uab{0.8} & \dar{0.9} & \uab{4.9} \\
\midrule
\multicolumn{6}{c}{\textbf{Experiment using control variable \cnontoxic{}}} \\
\midrule
\instruct & 0.31 & 0.15 & - & - & 29.8 \\
& \dab{29.0} & \dab{59.3} & & & \uab{7.8} \\
\bottomrule
\end{tabular}
}
\caption{Results for \threenine{} configuration on all the metrics for variations of \filter{} such as \filterFour{} and \filterThree{} in comparison with \instruct{}.}
\label{tab:filt-var}
\end{table}

\begin{table}[h]
\centering
\small{
\begin{tabular}{@{}l r r r r r}
\textbf{Model} &  \multicolumn{5}{c}{\textbf{96m-samples}}\\
\toprule
& EMT & TP & NLP & BD & E2E  \\
\cmidrule{2-6}
\base & 0.44 & 0.36 & 47.5 & 50.6 & 27.6 \\
\metadataEleven & 0.42 & 0.33 & 47.1 & 52.6 & 28.4 \\
& \dab{4.6} & \dab{8.8} & \dar{0.8} & \uab{4.1} & \uab{3.0} \\
\metadata & 0.41 & 0.31 & 48.1 & 50.1 & 28.5 \\
& \dab{5.9} & \dab{13.2} & \uab{1.4} & \dar{1.0} & \uab{3.2} \\
\metadataNinety & 0.42 & 0.33 & 47.0 & 47.6 & 28.6 \\
& \dab{2.9} & \dab{8.2} & \dar{1.1} & \dar{6} & \uab{3.8} \\
\instruct & 0.42 & 0.33 & 47.9 & 50.2 & 28.9 \\
& \dab{2.8} & \dab{6.8} & \uab{0.8} & \dar{0.9} & \uab{4.9} \\
\midrule
\multicolumn{6}{c}{\textbf{Experiment using control variable \cnontoxic{}}} \\
\midrule
\metadataEleven & 0.35 & 0.20 & - & - & 28.4 \\
& \dab{20.7} & \dab{43.5} & & & \uab{3.0} \\
\metadata & 0.33 & 0.18 & - & - & 28.3 \\
& \dab{24.0} & \dab{49.8} & & & \uab{2.6}  \\
\metadataNinety & 0.31 & 0.14 & - & - & 28.4 \\
& \dab{28.6} & \dab{59.9} & & & \uab{3.0} \\
\instruct & 0.31 & 0.15 & - & - & 29.8 \\
& \dab{29.0} & \dab{59.3} & & & \uab{7.8} \\
\bottomrule
\end{tabular}
}
\caption{Results for \threenine{} configuration on all the metrics for \metadata{} and \instruct{} in comparison with \metadataEleven{} and \metadataNinety{}.}
\label{tab:meda-var}
\end{table}

\begin{table*}[t]
\centering
\small{
\begin{tabular}{@{}l r r r r r @{\hskip 0.4in} r r r r r}
\textbf{Model} &  \multicolumn{5}{c}{\textbf{96m-samples}}{\hskip 0.4in} & \multicolumn{5}{c}{\textbf{150m-samples}} \\
\toprule
& EMT & TP & NLP & BD & E2E & EMT & TP & NLP & BD & E2E  \\
\midrule
\multicolumn{11}{c}{\textbf{Experiments with $\mathbf{357}$m parameter models}} \\
\midrule
\base & 0.44 & 0.36 & 47.5 & 50.6 & 27.6 & 0.43 & 0.35 & 48.2 & 50.0 & 30.8 \\
\instructeleven & 0.41 & 0.32 & 46.6 & 50.9 & 28.1 & 0.42 & 0.34 & 48.7 & 50.7 & 28.7 \\
& \dab{5.9} & \dab{11.8} & \dar{1.8} & \uab{0.5} & \uab{1.9} & \dab{2.0} & \dab{3.9} & \uab{1.0} & \uab{1.4} & \dar{6.8}\\
\instructfifty & 0.41 & 0.32 & 47.9 & 49.9 & 28.2 & 0.42 & 0.33 & 48.3 & 49.1 & 29.5 \\
& \dab{5.6} & \dab{11.9} & \uab{0.9} & \dar{1.5} & \uab{2.3} & \dab{3.0} & \dab{6.8} & \uab{0.2} & \dar{1.8} & \dar{4.3} \\
\instruct & 0.42 & 0.33 & 47.9 & 50.2 & 28.9 & 0.42 & 0.33 & 48.7 & 51.1 & 29.7 \\
& \dab{2.8} & \dab{6.8} & \uab{0.8} & \dar{0.9} & \uab{4.9} & \dab{1.9} & \dab{5.4} & \uab{0.9} & \uab{2.3} & \dar{3.7} \\
\midrule
\multicolumn{11}{c}{\textbf{Experiment using control variable \cnontoxic{} for $\mathbf{357}$m parameter models}} \\
\midrule
\instructeleven & 0.36 & 0.23 & - & - & 28.3 & 0.38 & 0.25 & - & - & 29.0 \\
& \dab{18.5} & \dab{36.8} & & & \uab{2.7} & \dab{13.0} & \dab{27.4} & & & \dar{6.0} \\
\instructfifty & 0.32 & 0.16 & - & - & 28.0 & 0.34 & 0.18 & - & - & 29.5 \\
& \dab{26.7} & \dab{54.6} & & & \uab{1.5} & \dab{22.3} & \dab{48.2} & & & \dar{4.4} \\
\instruct & 0.31 & 0.15 & - & - & 29.8 & 0.31 & 0.14 & - & - & 29.7 \\
& \dab{29.0} & \dab{59.3} & & & \uab{7.8} & \dab{28.1} & \dab{59.7} & & & \dar{3.5} \\
\midrule
\multicolumn{11}{c}{\textbf{Experiments with $1.3$b parameter model}} \\
\midrule
\base & 0.44 & 0.37 & 52.6 & 53.0 & 30.7 & 0.44 & 0.37 & 54.4 & 53.8 & 31.1 \\
\instructeleven & 0.41 & 0.32 & 52.9 & 54.1 & 33.5 & 0.42 & 0.33 & 54.3 & 54.0 & 31.0 \\
& \dab{6.3} & \dab{13.3} & \uab{0.5} & \uab{2.2} & \uab{9.1} & \dab{4.9} & \dab{11.0} & \dar{0.1} & \uab{0.3} & \dab{0.3} \\
\instructfifty & 0.41 & 0.32 & 53.4 & 54.5 & 30.8 & 0.42 & 0.34 & 53.9 & 54.6 & 32.6 \\
& \dab{6.3} & \dab{14.3} & \uab{1.4} & \uab{2.9} & \uab{0.4} & \dab{3.7} & \dab{8.4} & \dar{0.8} & \uab{1.4} & \uab{4.9} \\
\instruct & 0.43 & 0.34 & 53.3 & 53.9 & 30.6 & 0.42 & 0.34 & 53.7 & 54.9 & 31.7 \\
& \dab{3.6} & \dab{9.7} & \uab{1.3} & \uab{1.7} & \dar{0.2} & \dab{3.7} & \dab{9.2} & \dar{1.3} & \uab{2.0} & \uab{2.1} \\
\midrule
\multicolumn{11}{c}{\textbf{Experiment using control variable \cnontoxic{} for $\mathbf{1.3}$b parameter models}} \\
\midrule
\instructeleven & 0.32 & 0.15 & - & - & 34.2 & 0.33 & 0.18 & - & - & 32.1 \\
& \dab{28.3} & \dab{59.6} & & & \uab{11.7} & \dab{24.3} & \dab{51.4} & & & \uab{3.3}\\
\instructfifty & 0.31 & 0.14 & - & - & 30.9 & 0.32 & 0.15 & - & - & 32.9 \\
& \dab{29.6} & \dab{61.4} &  &  & \uab{0.7} & \dab{27.3} & \dab{58.4} & & & \uab{6.0} \\
\instruct & 0.31 & 0.15 & - & - & 31.3 & 0.31 & 0.14 & - & - &  32.6 \\
& \dab{30.2} & \dab{62.7} & & & \uab{2.1} & \dab{30.0} & \dab{63.5} & & & \uab{4.9} \\
\bottomrule
\end{tabular}
}
\caption{Results for $357$m and $1.3$b parameter  models on all the metrics for \instruct{} and its variations \instructeleven{} and \instructfifty{}.  For benchmark NLP tasks, bias detection tasks and E2E task we show the relative percentage improvement over \base{} with a \uab{} and decrement with a \dar{}. For the expected maximum toxicity and toxicity probability, we show the improvement with \dab{} because lower is better for these metrics.}
\label{tab:inst-var}
\end{table*}

\paragraph{Scaling the Model Size } 
Table~\ref{tab:8.3b-res} shows results for 8.3 billion parameter models which use \base{}, \filter{} and \instruct{} strategies on \threeone{} model configuration.

\paragraph{\filter{} Variations} Table~\ref{tab:filt-var} shows the results for \base{}, \filter{}, \filterFour{}, \filterThree{} and \instruct{} for all the \threenine{} model configuration on all eleven tasks.
These results are aggregated and presented in Fig.~\ref{fig:filt-var}.

\begin{figure}[!t]
\centering
{
\includegraphics[width=\linewidth]{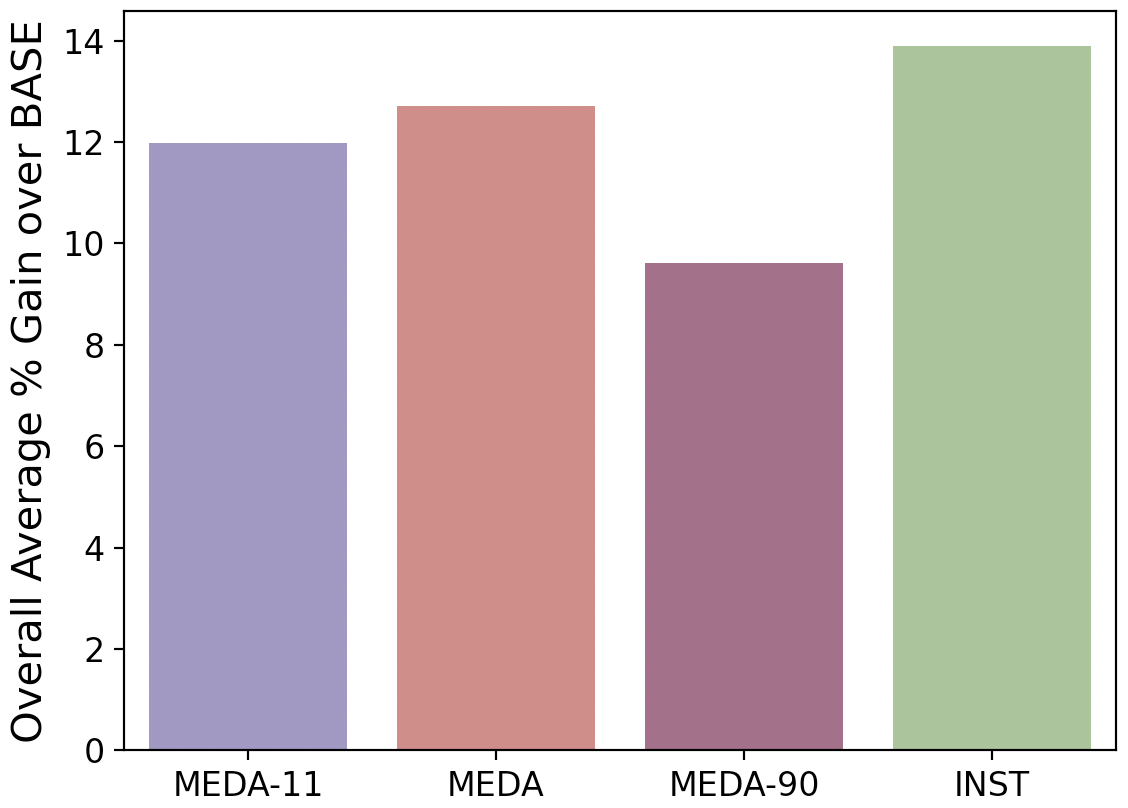}
}
\caption{Average the gains achieved by \metadataEleven{}, \metadata{}, \metadataNinety{}, and \instruct{} over \base{} across the eleven tasks for \threenine{} model configuration.}
\label{fig:meda-var}
\end{figure}

\paragraph{\metadata{} Variations} 
Table~\ref{tab:meda-var} shows the results for \base{}, \metadataEleven{}, \metadata{}, \metadataNinety{} and \instruct{} for \threenine{} model configuration on all eleven tasks.
These results are aggregated and presented in Fig.~\ref{fig:meda-var}.

\paragraph{\instruct{} Variations} Table~\ref{tab:inst-var} shows the results for \base{}, \instructeleven{}, \instructfifty{} and \instruct{} for all the four model configuration on all eleven tasks.
These results are aggregated and presented in Fig.~\ref{fig:inst-var}.



\end{document}